%% file: skeleton.tex
\documentclass[msc,deptreport,inf]{infthesis} 

\usepackage{amsmath}

\DeclareMathOperator*{\argmin}{arg\,min}
\usepackage{breqn}
\usepackage[ruled,linesnumbered]{algorithm2e}
\usepackage{cite}
\usepackage{wrapfig}
\usepackage{bbm}
\newtheorem{theorem}{Theorem}

\begin{document}
\begin{preliminary}

\title{On Training Implicit Meta-Learning With Applications to Inductive Weighing in Consistency Regularization}

\author{Fady Rezk}
\submityear{2021}

\abstract{
Meta-learning that uses implicit gradient have provided an exciting alternative to standard techniques which depend on the trajectory of the inner loop training. Implicit meta-learning (IML), however, require computing $2^{nd}$ order gradients, particularly the Hessian which is impractical to compute for modern deep learning models. Various approximations for the Hessian were proposed but a systematic comparison of their compute cost, stability, generalization of solution found and estimation accuracy were largely overlooked. In this study, we start by conducting a systematic comparative analysis of the various approximation methods and their effect when incorporated into IML training routines. We establish situations where catastrophic forgetting is exhibited in IML and explain their cause in terms of the inability of the approximations to estimate the curvature at convergence points. Sources of IML training instability are demonstrated and remedied. A detailed analysis of the effeciency of various inverse Hessian-vector product approximation methods is also provided. Subsequently, we use the insights gained to propose and evaluate a novel semi-supervised learning algorithm that learns to inductively weigh consistency regularization losses. We show how training a ``Confidence Network" to extract domain specific features can learn to up-weigh useful images and down-weigh out-of-distribution samples. Results outperform the baseline FixMatch performance.
}

\maketitle

\section*{Acknowledgements}
I would like to thank my supervisor, Professor Timothy Hospedales, for proposing and supervising this intriguing project. Your generosity with your time for supporting and mentoring me throughout the project was a great source of encouragement and learning. Your thought provoking questions that were asked at the perfect time have developed my skills and critical thinking as an aspiring researcher and I am grateful for that.

My sincere gratitude also goes to Dr. to be Ondrej Bohdal, my co-supervisor. Your prompt replies to my questions, making the time to meet me on short-notices even late in the evenings and spending generous amount of time discussing avenues of investigations was deeply appreciated. Whenever I lost focus or deviated from the main goals of my research, you have always helped me redirect my attention. In times of disorientation, your guidance fixated my focus again.

Professor Hospedales and Ondrej, your kindness when meetings ran over and generosity with your times to push this project to the current standard is deeply appreciated. Thanks for the great learning opportunity.

I would also like to thank Dr. Antreas Antoniou. Our discussions have always been a space of growing for me where I learnt to better read, critically evaluate and question research literature. Your readiness to always help and commitment to teaching has supported me throughout my Master's.

I would like to both thank and congratulate my close friend, Dr. Mohamed ElGhazaly, who only 2 days ago passed his PhD viva. Your check up calls have always been a wonderful source of support at times of stress. Your supportive and motivating words always helped me push myself. Our Pomodoro meetings were the funniest and most efficient study sessions I ever been in.

Many thanks for the generous support of Ahmed Ramy, CEO and founder of both TMentors Cairo and FastAutomate, for providing the GPU and compute resources used to run this dissertation's experiments.

Last but not least, I want to send out my gratitude and love to my family for their unwavering support during my whole Master's. Your support at times of stress and encouraging words were always heartwarming.
 \newpage\standarddeclaration

\tableofcontents
\end{preliminary}

\include{1.Introduction}
\include{2.Background}
\include{3.ImplicitMetaLearning}
\include{4.IML_Experiments}
\include{5.MetaFixMatch}
\include{6.Conclusion}
\bibliographystyle{unsrt}
\bibliography{0.references}

 \appendix
\include{99.appendix_perparamplots}
\include{99.appendix_ablation}
%
%

\end{document}

%% file: 1.Introduction.tex
\chapter{Introduction}
\section{Motivation}
With the advent of the internet and the ubiquity of data collection pipelines, enormous amounts of data have been and are being continuously curated, labeled, and annotated. Models trained on these datasets are considered ``intelligent" as they execute a variety of tasks after learning embedded patterns in the data, including but not limited to making predictions on new data as they arrive. Deep Learning \cite{goodfellow2016deep} has achieved remarkable performance in a wide variety of tasks such as image recognition \cite{726791,NIPS2012_c399862d}, natural language processing \cite{vaswani2017attention} and speech recognition \cite{6296526} among others.

Deep Learning, nonetheless, faces two major limitations. First, all deep learning algorithms require expensive hyperparameters optimization for improving performance. Secondly, deep learning models require a tremendous amount of labeled data to learn. With the time and monetary costs associated with manually annotating data especially in expert domains such as in medical research, it is paramount to capitalize on the availability of large unlabelled sets that might contain out-of-domain samples.

Semi-Supervised Learning (SSL) addresses the later limitation. Previous SSL approaches have achieved comparable and competitive results to supervised algorithms which train on a large scale labeled data \cite{fixmatch,kim2021selfmatch,xie2020unsupervised,berthelot2020remixmatch}. However, SSL algorithms introduce more hyperparameters to benefit from the unlabelled data. Additionally, it treats all unlabelled data as equally beneficial and attempting to disregard out-of-domain samples is as strenuous and laborious as labeling them.

Meta-Learning (or learning to learn) \cite{hospedales2020metalearning} paradigm allows to learn such meta-knowledge using same available data \cite{mpl} or with the cost of needing an extra small labelled dataset \cite{ren2020unlabeled}. Meta-Knowledge includes, but is not limited to, hyperparameters which address the first limitation of training Deep Learning models. Put simply, the meta-learning paradigm attempts to learn some knowledge about the learning process itself. The review \cite{hospedales2020metalearning} introduces a meta-learning taxonomy: Meta-Representation (what am I learning about the learning process?), Meta-Optimizer (How is this knowledge learned?) and Meta-Objective (Why is this knowledge learned?). The most famous Meta-Optimizers are gradient descent variants. MAML \cite{finn2017modelagnostic} is the most famous technique and by design is agnostic of the inner-loop task and model. Therefore, the method is easily applicable to any base model of choice and task which we aim to learn some meta-knowledge for.

Meta-Learning, however, requires tremendous memory and time resources. Therefore, gradient-based approaches such as MAML have focused on few-shot regimes. Recent research efforts have been oriented towards scaling Meta-Learning to many-shot regimes. Most famously, Implicit Meta-Learning (IML) \cite{lorraine2019optimizing} trains base model till convergence which allows using Implicit Function Theorem (IFT) to bypass unrolled differentiation and only backpropagates through the last step. It also claims to scale to millions of hyperparameters without placing extra strain on computational resources needed although using IFT introduces the requirement of computing second-order gradients and hence the need to provide estimates for the Hessian. The method offers an exciting alternative of another avenue of research since being able to use more inner-loop steps usually translates to finding better solutions in practice.

\section{Goals}
The concerns of this research project are two-fold. First, it is unclear how inverse hessian-vector product approximations, such as Neumann Series \cite{lorraine2019optimizing}, Conjugate Gradient or Identity Matrix approximation \cite{pmlr-v48-luketina16}, used in IML compare to the exact inverse Hessian. Previous studies have not scrutinized when these methods succeed, fail or when to choose one over the other. Training the inner-loop till convergence is usually ignored in practice and IML is used in semi-online settings (taking many inner steps without reaching convergence before a meta-update) without investigating how this influences the performance of overall solutions. In addition, there is a knowledge gap in the research literature for systematically studying the costs of respective approximations or systematically juxtaposing the performance of solutions found by each method. Therefore, the first part of the study introduces a comparative analysis of different inverse Hessian approximation techniques and their influence on the accuracy of the solution found, memory consumption, compute cost, comparison to exact inverse hessian, and their respective stability. Additionally, the convergence assumption is revisited.\\
Subsequently, insights are used to design a novel inductive semi-supervised algorithm that learns to weigh unlabelled data points in a meta-learning setting. Our algorithm learns to learn prioritizing unlabelled images. To maximize base model performance, it learns to down-weigh out-of-domain samples and emphasizes relevant ones. This is achieved through building meta-knowledge regarding which visual features directly contribute to improving semi-supervised algorithm accuracy and loss. This meta-visual feature extractor is denoted Confidence Network. The Confidence Network's learned representation transforms unlabelled images to a weight resembling how \textit{confident} is the network that the unlabelled image would improve the base models' performance.\\
\textbf{In summary, the goals of the project are:}
\begin{enumerate}
    \item Investigate Implicit Meta-Learning training, compare inverse Hessian-vector product approximation methods estimation accuracy against exact computation and juxtapose the methods in terms of generalization of the solution found, compute cost, and stability
    \item Design and evaluate a novel inductive semi-supervised learning algorithm that exploits the power of Implicit Meta-Learning
\end{enumerate}
\section{Contributions}
The contributions of this study are:
\begin{enumerate}
    \item Present how catastrophic forgetting phenomenon is exhibited in IML, establish one source/cause of it in terms of approximation method inability to estimate the loss function curvature beyond convergence, and propose a basic solution to it
    \item Show and explain the counter-intuitive fact that using more terms/iterations in either Neumann Series or Conjugate Gradient approximations degrades overall approximation power and final solution performance
    \item Establish sources of instability in IML, namely under-estimation of training loss at the start of training, and propose a solution that both remedies the issue and cuts down the base model training steps till convergence by half
    \item Show the power of IML, if trained correctly, in significantly improving over baselines
    \item Provide a detailed evaluation of the efficiency and estimation accuracy of different $2^{nd}$ order gradient approximation methods employed in implicit meta-learning
    \item Propose and evaluate a novel semi-supervised algorithm that outperforms FixMatch baseline by learning to inductively weigh consistency regularization loss
    \item Establish shortcomings of our proposed method and propose future directions to resolve and address them
\end{enumerate}
\section{Report Outline}
In chapter \ref{chapter:background}, the necessary and required background to understand the dissertation is covered. Relevant literature is reviewed to ground this dissertation's investigations and proposals in previous work. Next, chapter \ref{chapter:iml} provides a detailed and formal treatment of Implicit Meta-Learning along with the various inverse-Hessian approximation methods employed in the comparative analysis. Subsequently, the experiments conducted, results, findings, and discussions for the first goal of this dissertation are detailed in chapter \ref{chapter:phase1}.\\
The findings and insights of the first part of this dissertation are then used to design a novel Semi-Supervised Learning algorithm. The algorithm, title Fixing FixMatch, is introduced and derived formally in chapter \ref{chapter:fixingfixmatch}. The training details, preliminary experiments, and initial results are also covered besides a critique of the proposal and directions for further improvement. Finally, chapter \ref{chapter:conclusion} concludes the dissertation and summarizes all the work done herein.

%% file: 2.Background.tex
\chapter{Background}\label{chapter:background}
\section{Meta-Learning Overview}
This study focuses on gradient-based Meta-Learning which is the most efficient and widely adopted formulation. For a full review on Meta-Learning, please refer to the most recent survey by \textit{Hospedales et al.} \cite{hospedales2020metalearning}. To introduce Meta-Learning, one has to start with Model Agnostic Meta-Learning (MAML) \cite{finn2017modelagnostic}. MAML, as said earlier, focuses on the few-shot regime and was designed with few-shot learning in mind. It attempts to learn some meta-knowledge such as deep network initialization $\theta=\theta_0$ whose meta-objective is generalizing to new unseen downstream tasks as fast as possible with few labels per class. MAML is a great starting point to introduce Meta-Learning.

Let $f_\theta$ be a deep network parameterized by $\theta$. Given a distribution over tasks $p(\mathcal{T})$, MAML samples a task $\mathcal{T}_i$ and trains the base model on it using a support (training) set for few episodes. Subsequently, a query (validation) set is used to evaluate the performance of the base model to update the meta-knowledge. MAML produces new base model parameters $\theta_i'$ by taking one or multiple standard gradient-descent steps, i.e: $\theta_i'=\theta-\alpha\nabla_\theta\mathcal{L}_{\mathcal{T}_i}(f_\theta)$ where $\alpha$ is the learning rate (can be fixed or meta-learned). Thereafter, the meta-knowledge $\theta$ is updated by minimizing the validation loss on the query set across all sampled tasks after few-shot training:
\begin{equation}\label{eq:maml}
   \argmin_\theta \sum_{\mathcal{T}_i\sim p(\mathcal{T}_i)}\mathcal{L}_{\mathcal{T}_i}(f_{\theta_i^{'}}) = \sum_{\mathcal{T}_i\sim p(\mathcal{T}_i)}\mathcal{L}_{\mathcal{T}_i}(f_{\theta_i^{'}=\theta-\alpha \nabla_\theta \mathcal{L}_{\mathcal{T}_i}(f_\theta)})
\end{equation}
Please note that equation \ref{eq:maml} is a bi-level optimization problem. This is the standard setup for all gradient-descent based meta-learning algorithms. The training (support) and validation (query) datasets samples should not overlap.

MAML was found to be unstable, sensitive to deep architecture choices, and requires heavy tuning of hyper-parameters \cite{antoniou2019train}. Additionally, extending MAML to many-shot regimes is impractical. Backpropagating through many inner loop steps suffers from vanishing gradients and stresses memory for the need to store intermediate steps gradients.

Various approaches were introduced to address the shortcomings of MAML. These include Forward-Mode Differentiation (FMD) which provides exact solutions and does not require the inner loop loss to be a function of the meta-knowledge optimized for \cite{pmlr-v70-franceschi17a,micaelli2021nongreedy}. FMD is, however, not scalable with the meta-knowledge. Moreover, greedy online methods are another alternative \cite{liu2018darts,gunes2018online} but they suffer from short-horizon bias \cite{wu2018understanding}. The most promising technique is using implicit gradients \cite{rajeswaran2019metalearning,lorraine2019optimizing}. They require the inner loop loss to be a function of the meta-knowledge. Therefore, they are mostly used in hyperparameters optimization which is our concern in this study. For a wider review, please refer to \textit{Hospedales et. al, 2020} survey \cite{hospedales2020metalearning}. Implicit Meta-Learning is introduced in detail in chapter \ref{chapter:iml}.
\section{Semi-Supervised Learning}
\subsection{Overview}
Semi-Supervised Learning (SSL) learns from both a labelled dataset $\mathcal{D}=\{(x,y)\}$ and unlabelled dataset of only input instance $\mathcal{U}=\{(x)\}$. SSL objective function is divided into two terms for each dataset:
\begin{equation}\label{eq:ssl}
    \min_\theta \sum_{(x,y)\in\mathcal{D}}l_S(x,y,\theta)+\lambda\sum_{u\in\mathcal{U}}l_U(u,\theta)
\end{equation}
where $l_S$ is the standard supervised loss such as cross-entropy for classification, $l_U$ is an unsupervised loss on the unlabelled dataset and $\lambda$ is a hyperparameter. The unsupervised loss chosen gives rise to different SSL algorithms. Here, the focus is on SSL algorithms that benefited from Meta-Learning; especially Consistency Regularization (CR) and Proxy-Label (PL) Methods. For a full review, please to refer to \cite{ouali2020overview}. More algorithms are also covered in this dissertation's proposal.
\subsubsection{Proxy-Label Methods}
First, PL methods use the unlabelled dataset by generating pseudo-labels $\tilde{y}$ for each unlabelled instance by some heuristic. Most commonly a teacher network pre-trained on a large labeled dataset such as ImageNet is used or a network trained on the available small annotated dataset. The pseudo-labels along with the small labelled dataset are in turn used together to train a student network using a supervised loss, i.e: $l_U(u,\theta)\equiv l_S(u,\tilde{y},\theta)$. This approach was found to suffer from confirmation bias \cite{9207304}. Recently, it was proposed to introduce uncertainty weight per each unlabelled sample in a transductive setting through a regularization term that motivates compactness and separation between classes \cite{Shi_2018_ECCV}.

\subsubsection{Consistency Regularization}
CR capitalizes on the unlabelled data using a different approach. It assumes that a base network's output should not change significantly if the unlabelled sample is perturbed or augmented. Unsupervised Data Augmentation, for example, uses AutoAugment and Back Translation \cite{xie2020unsupervised}. The base network takes both the unlabelled sample and its augmented version as inputs. The unaugmented instance output is used as a pseudo-label. The unsupervised loss in equation \ref{eq:ssl} becomes the cross-entropy between the pseudo label and the soft labeled output of the augmented image. This loss regularizes the network to produce similar outputs to both original and augmented unlabelled instances. Hence, the network benefits from implicit information in the unlabelled dataset when learning. Additionally, CR techniques include FixMatch \cite{fixmatch}, MixMatch \cite{mixmatch}, ReMixMatch \cite{berthelot2020remixmatch}, Virtual Adversarial Training (VAR) \cite{miyato2018virtual} which all use the same assumption but slightly different augmentation strategies.

\subsection{FixMatch}
In this study, we use FixMatch as our base semi-supervised algorithm \cite{fixmatch} whose backbone model is a deep network, $f_\theta$, parameterized by $\theta$. Therefore, the algorithm is explained briefly in this section as the necessary background for the rest of the report. Put simply, FixMatch uses two data augmentation strategies for the unlabelled instances. The two strategies are divided into a weak augmentation $\alpha(u)$ and a strong one $\mathcal{A}(u)$. The weak strategy is flipping an image horizontally with a probability of 50\% and translating the image by up to 12.5\% vertically or horizontally. Meanwhile, the strong augmentations used and evaluated were RandAugment \cite{NEURIPS2020_d85b63ef} and CTAugment \cite{berthelot2020remixmatch}. In this study, we only use RandAugment to simplify implementation and supplement initial comparisons with the baseline.
\begin{figure}[tb]
\begin{center}
\centerline{\includegraphics[width=0.75\columnwidth]{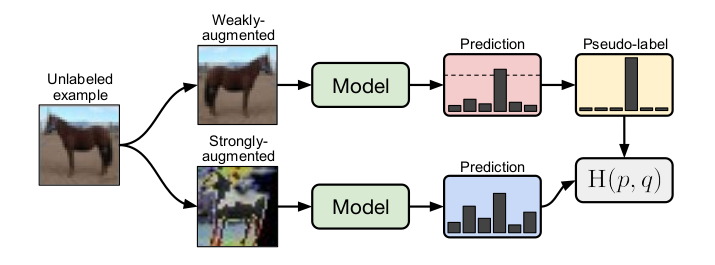}}
\caption{FixMatch Unsupervised Loss. \cite{fixmatch}}
\label{fig:fixmatch_working}
\end{center}
\end{figure}

As shown in figure \ref{fig:fixmatch_working}, each unlabelled image is augmented twice using both weak and strong perturbations. Let $\alpha(u)$ denote the weak augmentation function on instance $u$, then the prediction, $q=f_\theta(y|\alpha(u))$, is used to produce a pseudo-label, $\hat{q}=argmax(q)$, if the highest predicted class exceeds a threshold $\tau$; weak augmented images predictions are masked to only include confidently predicted instances. Subsequently, the unsupervised loss is the cross-entropy between the pseudo-label and the soft labelled output of the strong augmented image (regularizing the network to predict similar outputs to both strong and weak augmentations). The unsupervised loss per unlabelled image is then:
\begin{equation}
    l_U(u)=\mathbbm{1}(max(q)\geq\tau)\text{H}(\hat{q}, f_\theta(y|\mathcal{A}(u)))
\end{equation}
where $\text{H}(.,.)$ is the standard cross-entropy classification loss and $\mathcal{A}(u)$ is the strong augmentation on unlabelled image $u$. In summary, FixMatch combines both pseudo-labels and consistency regularization by design.

\section{Meta-Learning for SSL}
Meta Psuedo Labels (MPL) \cite{mpl} was introduced to fix the confirmation bias of PL methods using the bi-level optimization formulation of Meta-Learning. MPL trains both the teacher and student network alternately to improve each other's performance. This alternating setup is mutually beneficial for both networks. First, the teacher generates pseudo labels which are used to train the student. Subsequently, the performance of the student is used as a signal to update the teacher parameters in the direction of the solution space which improves the student performance. However, producing pseudo-labels for all unlabelled images assumes that they are all equally beneficial with not out-of-domain samples.

Regarding Consistency Regularization, by referring to equation \ref{eq:ssl}, one would note that all unlabelled instances are weighted equally by the hyperparameter $\lambda$. If unlabelled instances contain out-of-domain samples, weighting them equally would harm performance. Therefore, \textit{Ren et al} proposed to learn a weight per each unlabelled image transductively \cite{ren2020unlabeled}.

\section{Critique}
To the best of the author's knowledge, no study has previously attempted to learn to weigh unlabelled datapoints inductively for consistency regularization by extracting domain-centric visual features from the unlabelled images. Meta Pseudo Labels learns to label instances by assuming that they are equally beneficial. Furthermore, \textit{Ren et.al} learn instance wise weights transductively \cite{ren2020unlabeled}. Similarly, in a self-supervised pre-training setting, \textit{Ericsson et al.} learns to prune data transductively to minimize downstream tasks loss of a different domain \cite{ericsson2020dont}. The proposed method learns a probability distribution over the weight of each unlabelled instance.

%% file: 3.ImplicitMetaLearning.tex
\chapter{Implicit Meta-Learning}\label{chapter:iml}
\section{Overview}
Implicit Meta-Learning (IML) has found great interest and success stories in recent literature on meta-learning for many-shot regimes. In particular, \textit{Lorraine et al.} \cite{lorraine2019optimizing} have introduced Implicit Function Theorem to the meta-update rule. This allows the inner loop to train to convergence without having to unroll the trajectory of the inner-loop and backpropagate through all of this. First, the method is introduced.

In hyperparameters Optimization (HO), IML provides approximate gradients for the long inner loop trajectory of the baseline model training process. HO as a bi-level optimization problem is defined as:
\begin{equation}
    \boldsymbol\lambda^*=\argmin_\lambda \mathcal{L}_V^*(\boldsymbol\lambda), \textrm{where}    
\end{equation}
\begin{equation}
    \mathcal{L}_V^*(\boldsymbol\lambda):=\mathcal{L}_V(\boldsymbol\lambda,\textbf{w}^*(\boldsymbol\lambda)) \textrm{ and } \textbf{w}^*(\boldsymbol\lambda):=\argmin_{\textbf{w}}\mathcal{L}_T(\boldsymbol\lambda,\textbf{w})
\end{equation}
where $\mathcal{L}_T$ and $\mathcal{L}_V$ are training and validation losses respectively. The variables $\lambda$ are the hyperparameters being optimized for and $\textbf{w}^*(\boldsymbol\lambda)$ are base model best-response parameters to the hyerparameters; that is training the base model till convergence given the hyperparameters. Using the chain rule to take the derivative of the validation loss with respect to (w.r.t) the hyperparameters, we get:
\begin{equation}\label{eq:dvdlambda}
        \frac{\partial\mathcal{L}_V^*}{\partial\boldsymbol\lambda}=\left(\frac{\partial\mathcal{L}_V}{\partial\boldsymbol\lambda}
    +\frac{\partial\mathcal{L}_V}{\partial\textbf{w}}
    \frac{\partial\textbf{w}^*}{\partial\boldsymbol\lambda}
    \right)\Biggr|_{\boldsymbol\lambda,\textbf{w}^*(\boldsymbol\lambda)}=\left(\frac{\partial\mathcal{L}_V(\lambda,\textbf{w}^*(\boldsymbol\lambda))}{\partial\boldsymbol\lambda}
    +\frac{\partial\mathcal{L}_V(\lambda,\textbf{w}^*(\boldsymbol\lambda))}{\partial\textbf{w}^*(\boldsymbol\lambda)}
    \frac{\partial\textbf{w}^*(\boldsymbol\lambda)}{\partial\boldsymbol\lambda}
    \right)
\end{equation}
The three terms in equation \ref{eq:dvdlambda} from left to right are: hyperparameters direct gradient, base model parameters direct gradient, and best-response Jacobian. In HO, the first term is usually zero since the hyperparameters are not used during inference time (on the validation set). The second term is easily computable using automatic differentiation library as a standard routine in all machine learning algorithms. Finally, the Best-Response Jacobian provides information on how the weights have evolved and changed with respect to the hyperparameters used. It requires unrolled differentiation which as discussed earlier is problematic. Therefore, Implicit Function Theorem (IFT) is used in Lorraine's approach.

\begin{theorem}\label{theorem:ift}
(Implicit Function Theorem). Given regularity conditions, if $\frac{\partial\mathcal{L}_T}{\partial\textbf{w}}|_{\boldsymbol\lambda',\textbf{w}'}=0$ for some values $(\boldsymbol\lambda',\textbf{w}')$, then $\exists\textbf{w}^*(\boldsymbol\lambda)$ s.t. $\frac{\partial\mathcal{L}_T}{\partial\textbf{w}}|_{\boldsymbol\lambda,\textbf{w}^*(\boldsymbol\lambda)}=0$ and
\begin{equation}\label{eq:ift}
        \frac{\partial\textbf{w}^*}{\partial\boldsymbol\lambda}\Biggr|_{\boldsymbol\lambda^{'}}=-\left[\frac{\partial^2\mathcal{L}_T}{\partial\textbf{w}\partial\textbf{w}^T}\right]^{-1}\times\frac{\partial^2\mathcal{L}_T}{\partial\textbf{w}\partial\boldsymbol\lambda^T}\Biggr|_{\boldsymbol\lambda^{'},\textbf{w}^*(\boldsymbol\lambda^{'})}
\end{equation}
\end{theorem}
The condition of theorem \ref{theorem:ift} is equivalent to a fixed point in the base model training gradient field. If the base model is trained till convergence, then the condition hold, and IFT is used to avoid unrolled differentiation and find a local approximation to the best-response Jacobian. Implicit Meta-Learning is shown in algorithm \ref{alg:iml}.

\begin{algorithm}[H]\label{alg:iml}
\caption{Implicit Meta-Learning Algorithm}
\SetAlgoLined
\SetKwInOut{Input}{Input}
\SetKwInOut{Output}{Output}
\Input{Training Data $\mathcal{D}$, Validation Data $\mathcal{V}$.}
\Output{Optimal Hyperparameters $\boldsymbol\lambda^*$}
Initialize hyperparameters and base model parameters as $\boldsymbol\lambda$ and $\textbf{w}$ respectively

\While{\text{Not Converged}}{
    \For{t = 0...T-1}{
        \textbf{w}' -= $\alpha . \frac{\partial\mathcal{L}_T}{\partial\textbf{w}}|_{\boldsymbol\lambda',\textbf{w}'}$\\
    }
    $v_1=\frac{\partial\mathcal{L}_V(\boldsymbol\lambda, \textbf{w}^*(\boldsymbol\lambda^{(t)}))}{\textbf{w}^*(\boldsymbol\lambda^{(t)})}$
    
    Approximate inverse Hessian-vector product, $v_2=v_1\times\left[\frac{\partial^2\mathcal{L}_T}{\partial\textbf{w}\partial\textbf{w}^T}\right]^{-1}$
    
    $v_3$= $v_2\times\frac{\partial^2\mathcal{L}_T}{\partial \textbf{w}\partial\boldsymbol\lambda}$

    
    $\boldsymbol\lambda^{(t+1)}-=\eta\times (-v_3) \rightarrow \boldsymbol\lambda^{(t+1)}+=\eta\times v_3$
    
}
\end{algorithm}
The algorithm involves computing the Inverse Hessian-vector product of the base model as per line 7. For deep neural networks, this computation is practically impossible memory-wise and is extremely computationally expensive. In the coming section, commonly used approximation methods for inverse Hessian-vector products are introduced and discussed while showing how they give rise to different algorithms.

\section{Approximating Inverse Hessians}
\subsection{Neumann Series}
Lorraine's approach used Neumann Series to bypass the need to invert the Hessian matrix which requires cubic operations to the size of the matrix $\mathcal{O}(n^3)$. The Neumann Series for an inverse Hessian is defined as:
\begin{equation}
    \left[\frac{\partial^2\mathcal{L}_T}{\partial\textbf{w}\partial\textbf{w}^T}\right]^{-1} = \lim_{i \to \infty}\sum_{j=0}^i\left[I-\frac{\partial^2\mathcal{L}_T}{\partial\textbf{w}\partial\textbf{w}^T}\right]^j
\end{equation}
The paper further shows that approximating this inverse using $i$ terms in the Neumann Series is equivalent to unrolling differentiation for a trajectory of $i$ steps around the local optimal weights of the base model. This theorem is re-stated in Theorem \ref{theorem:unrolledift}. They further show that storing the inverse Hessian in memory which is impractical can be detoured by utilizing an efficient vector-Hessian product. Theorem \ref{theorem:unrolledift} gives rise to algorithm \ref{alg:imlift}. The arxiv paper has a typo in line 4 ($p-=v$). The correct computation is $p+=v$.
\begin{theorem}\label{theorem:unrolledift}
(Neumann-SGD). If we recurrently unroll Stochastic Gradient Descent (SGD) starting from $\textbf{w}_0=\textbf{w}^*(\lambda)$, then
\begin{equation}
    \frac{\textbf{w}_{i+1}}{\partial\boldsymbol\lambda}=\left(\sum_{j<i}\left[I-\frac{\partial^2\mathcal{L}_T}{\partial\textbf{w}\partial\textbf{w}^T}\right]^j\right)\frac{\partial^2\mathcal{L}_T}{\partial\textbf{w}\partial\boldsymbol\lambda}\Biggr|_{\textbf{w}^*(\boldsymbol\lambda)}
\end{equation}
and if $I+\frac{\partial^2\mathcal{L}_T}{\partial\textbf{w}\partial\textbf{w}^T}$ is contractive, then as $i\to\infty$, the reccurence converges to eq. \ref{eq:ift}.
\end{theorem}
\begin{algorithm}[H]\label{alg:imlift}
\caption{Neumann Series Approximation of the Inverse Hessian-vector Product}

\SetAlgoLined
\SetKwInOut{Input}{Input}
\SetKwInOut{Output}{Output}
\Input{Direct Validation Loss derivative w.r.t base model weights \textbf{v}, direct Training Loss derivative w.r.t base model weights \textbf{f}, base model weights $\textbf{w}$ and number of neumann steps $i$}
\Output{Inverse Hessian Vector Product, \textbf{p}}
Initialization: Store copy of validation loss direct gradient \textbf{p} = \textbf{v}\\
    \For{j = 1...i}{
    \textbf{v}$-=\alpha\times$ grad(\textbf{f}, $\textbf{w}$,grad\_outputs=\textbf{v})\\
    \textbf{p}+=\textbf{v}
}
\end{algorithm}

\subsection{Conjugate Gradient}
Conjugate Gradient (CG) is a fabled algorithm for Hessian-free second order optimization \cite{1448859,Martens2012}. The implemented version of CG in this study is based on lorraine's original Github implementation from their Data Augmentation experiments. In meta-learning terms, CG aims to approximate the inverse Hessian-vector product by solving the following equation:
\begin{equation}
    \argmin_{\textbf{x}}||\textbf{x}\frac{\partial^2\mathcal{L}_T}{\partial\textbf{w}\partial\textbf{w}^T}-\frac{\mathcal{L}_V}{\partial\textbf{w}}||
\end{equation}
CG originally was derived from Newton's optimization method which solves an optimization problem through solving a sequence of local quadratic approximations of the function to update model parameters. Meanwhile, CG for conducing Hessian-Free Deep Learning optimization requires damping and pre-conditioning \cite{10.5555/3104322.3104416}. To minimize the degrees of freedom for CG algorithm and achieve comparable outcomes, Lorraine's implementation was depended on as said. The hyperparameters update rule is then:
\begin{equation}\label{eq:cg}
    \lambda^{(t+1)} = \boldsymbol\lambda^{(t)} +\eta
    \left(
    \argmin_{\textbf{x}}\Biggr|\Biggr|\textbf{x}\frac{\partial^2\mathcal{L}_T}{\partial\textbf{w}\partial\textbf{w}^T}-\frac{\partial\mathcal{L}_V(\lambda^{(t)},\textbf{w}(\boldsymbol\lambda))}{\partial\textbf{w}(\boldsymbol\lambda)}\Biggr|\Biggr|
    \right)
    \frac{\partial^2\mathcal{L}_T}{\partial\textbf{w}\partial\boldsymbol\lambda^T}\Biggr|_{\boldsymbol\lambda^{(t)},\textbf{w}^*(\boldsymbol\lambda^{(t)})}
\end{equation}

\subsection{Identity Matrix For IML vs T1-T2}
Finally, using the Identity matrix to approximate the Inverse Hessian gives rise to T1-T2 algorithm \cite{pmlr-v48-luketina16}. T1-T2 is a greedy algorithm. It updates the hyerparameters being optimized for after every single update for the base model parameters. Using the identity matrix simply means that the derivative of the weight with respect to the gradient is approximated as:
\begin{equation}
    \frac{\partial\textbf{w}}{\partial\boldsymbol\lambda}\Biggr|_{\boldsymbol\lambda^{'}}=-\frac{\partial^2\mathcal{L}_T}{\partial\textbf{w}\partial\boldsymbol\lambda^T}\Biggr|_{\boldsymbol\lambda^{'},\textbf{w}(\boldsymbol\lambda^{'})}    
\end{equation}
Subsequently, the hypergradient using the following rule:
\begin{equation}\label{eq:t1t2}
    \lambda^{(t+1)} = \boldsymbol\lambda^{(t)} +\eta\frac{\partial\mathcal{L}_V(\lambda^{(t)},\textbf{w}(\boldsymbol\lambda))}{\partial\textbf{w}(\boldsymbol\lambda)}\frac{\partial^2\mathcal{L}_T}{\partial\textbf{w}\partial\boldsymbol\lambda^T}\Biggr|_{\boldsymbol\lambda^{(t)},\textbf{w}(\boldsymbol\lambda^{(t)})}
\end{equation}
The assumption of T1-T2 is that equation \ref{eq:t1t2} approximates the hypergradient and the identity approximates the true Inverse Hessian. The method theoretically only converges when all components of the best-response Jacobian and mixed partial derivative are orthogonal. Therefore, there are no guarantees of convergence since machine learning optimization converges when gradients are theoretically zero. In theory, this approximation is unjustified and will not strictly hold in practice. Nevertheless, in experiments where the base model is ResNet \cite{7780459}, for example, one might argue that the residual connections do bring the Hessian closer to the Identity.

An important distinction in terminology for the rest of the report is T1-T2 vs IFT-Identity (or simply referred to as Identity approximation). Whenever T1-T2 is used, then the algorithm alternates between the base model and meta-knowledge updates each step (one meta-update after every base model update). Otherwise, IFT-Identity is used when the base model is trained for many steps before taking a meta-update step whose hyper-gradient is computed by using the identity in place of the Inverse Hessian.
\subsection{Note}
Please note that discussed algorithms above are practically realizable as long as they depend on Hessian-Vector products. These are assumed to be easily computable without storing the actual Hessian in memory through Pearlmutter's method \cite{pearlmutter}.

\section{Proposed Methodology}
As stated earlier, it is unclear how different inverse-Hessian approximation methods influences and affects implicit meta-learning. Previous studies resort to comparisons of the approximations to the true inverse on over-simplified problems such as linear networks \cite{lorraine2019optimizing} or some fixed chosen random matrices. Otherwise, approximation methods accuracy with respect to the exact inverse Hessian of deep learning models, when studied in more depth, would only involve simplified tractable sub-problems such as only computing the exact inverse for the final layer of a deep ResNet \cite{ren2020unlabeled}. It is unclear how these methods influence and affect the overall performance of solutions found when incorporated into the training procedure itself or how the approximations compare to the exact inverse as training proceeds and the loss landscape is being explored. Equally important, the overall cost of training using the approximations and their stability has been utterly overlooked and ignored in the literature.\\
We propose a comparative analysis between Neumann Series, Conjugate Gradient, IFT-Identity, and T1-T2 approximations to contrast empirically the pros and cons of each method. We admit a small model. A Multi-Layer Perceptron (MLP) with a single hidden layer and as many hidden neurons as inputs is chosen. Using MLP is selected to minimize the complexity of the problem, remove the influence of strong architecture prior inductive biases, and enable computing exact inverse Hessians.\\
The meta-knowledges learnt are: 1) Per Parameter $L_2$ Regularizer and 2) Per Layer $L_2$ Regularizer. Meta-Knowledge (1) is an over-parameterization of the base model regularization. This is used to question whether using different approximations successfully overfits the validation (query) set. Overfitting the validation set translates to IML making the most out of available information in the validation data. Meta-knowledge (2) is a standard regularization technique that is used to improve base model generalization power by balancing the trade-off between learning from the validation set and improving base model generalization.\\
The hyperparameters were initialized randomly for both meta-knowledge. The weight decay values are kept positive by applying a Softplus non-linearity to them. The total regularization value is computed as: $\left[\text{Softplus}(\boldsymbol\lambda)\odot \textbf{w}\right]^2$ where $\odot$ is a point-wise multiplication. The hyperparameters $\boldsymbol\lambda$ are either a matrix with the same shape as an MLP layer's parameters $w$ (meta-knowledge 1) or a single element for each layer parameter's (meta-knowledge 2).\\
The metrics colligated on meta-knowledge (1) were 1) approximation accuracy with respect to exact inverse, 2) stability of the training procedure, 3) approximator influence on training procedure capacity (ability to overfit a small validation set), and 4) compute cost in terms of memory and time. For the memory metric, the maximum allocated memory for computing the inverse-Hessian vector product is measures. Meta-Knowledge (2) is used to test the generalization power of the solution found, algorithm stability, and again on compute cost. Each experiment setup was repeated three times on two datasets, namely MNIST \cite{deng2012mnist} and CIFAR-10 \cite{cifar} where the latter is a more challenging dataset than the former to learn using an MLP.\\

%% file: 4.IML_Experiments.tex
\chapter{Experiments and Results}\label{chapter:phase1}
For all experiments of this part of the project, the base model optimizer used was Adam \cite{kingma2017adam} with a learning rate (LR) of $10^{-4}$ and RMSprop as the meta-knowledge optimizer with LR of 0.1 and default Pytorch values for all other parameters. Experiments were run on a single Tesla k80 GPU and each was repeated three times with different seeds.
\section{Learning Per Parameter Regularizer}
\subsection{Training Details}
For this problem setup, 50 samples were used per training, validation, and testing set. The inner loop was trained for 50 steps with batch sizes of 32 and meta-knowledge was updated 1000 times; 1000 epochs of 50 steps each, updating every epoch. For T1-T2, the model was trained for a total of 50K steps updating meta-knowledge after every step. The number of inner steps and batch size allow the base model performance to converge. Using few samples for training/validation enables investigating the power of approximation methods in using the validation set as stated earlier. This experiment follows directly from Lorraine's investigation (\cite{lorraine2019optimizing}, section 5.1). The number of Neumann Series terms studied were 3, 10, 20, 35, and 50. For CG, only 3, 5, and 10 steps were used because it has a heavy compute time cost as will be shown briefly.
\subsection{Overfitting a Small Validation Set Results}
The results are shown at table \ref{tab:perparam_perfmnist} and \ref{tab:perparam_perfcifar} for MNIST and CIFAR-10 datasets respectively. For Brevity, Neumann (3) and CG (3) will be used to respectively denote Neumann Series or CG approximations using 3 terms. It was found that if IML is trained for long after convergence, then the validation and training accuracies start degrading. This behavior did not manifest in T1-T2. These dynamics are shown in appendix \ref{app:overfitting_detailed} along with more detailed experiment results. Therefore, the reported validation and training accuracies are the highest values achieved during the whole training procedure before degradation. Shortly, early stopping will be investigated as means for mitigating this issue. For comparison, results without and with early stopping are shown in columns ``Test" and ``ES Test". ``Test" accuracy is that of the final model after full training (at the end of 1000th epoch), while ``ES Test" is evaluated on the model checkpoint of early stopping (model at the point where validation loss was at its lowest during the whole training before early stopping termination).\\
\begin{table}
\centering
\caption{\textbf{MNIST:} Mean and standard deviation of performance metrics for learning per parameter weight decay regularizer (overfitting small validation set) experiment. ``ES Test" and ``Test" are results of final model performances with and without early stopping. The increasing number of Neumann or CG steps produces instabilities. Neumann (3) is best performing method with IFT-Identity producing comparative results with less variance.}
\label{tab:perparam_perfmnist}
\begin{tabular}{lllll}
\hline
  & \multicolumn{4}{c}{Accuracy}  \\ 
\cline{2-5}
Name   &  Training   & Validation & Test & ES Test\\
\hline
3 Step Neumann&100.00$\pm$0.00\%&100.00$\pm$0.00\%&64.00$\pm$2.83&77.33$\pm$4.11\\
10 Step Neumann&100.00$\pm$0.00\%&88.00$\pm$5.66\%&48.00$\pm$10.20&54.67$\pm$6.18\\
20 Step Neumann&100.00$\pm$0.00\%&79.33$\pm$0.94\%&56.00$\pm$10.71&60.00$\pm$6.53\\
35 Step Neumann&100.00$\pm$0.00\%&91.33$\pm$5.73\%&52.00$\pm$4.32&60.00$\pm$3.27\\
50 Step Neumann&98.00$\pm$2.83\%&81.33$\pm$10.50\%&37.33$\pm$19.48&56.00$\pm$5.89\\
3 Step CG&100.00$\pm$0.00\%&94.67$\pm$0.94\%&57.33$\pm$1.89&57.33$\pm$1.89\\
5 Step CG&99.33$\pm$0.94\%&78.00$\pm$4.32\%&62.00$\pm$3.27&52.67$\pm$0.94\\
10 Step CG&58.67$\pm$0.94\%&40.00$\pm$2.83\%&12.00$\pm$0.00&6.00$\pm$0.00\\
IFT-Identity&100.00$\pm$0.00\%&94.00$\pm$1.63\%&72.00$\pm$4.32&76.67$\pm$1.89\\
T1-T2&100.00$\pm$0.00\%&96.00$\pm$1.63\%&69.33$\pm$2.49&-\\
\hline
\end{tabular}
\end{table}
\begin{table}
\centering
\caption{\textbf{CIFAR-10:} Mean and standard deviation of performance metrics for learning per layer weight decay regularizer experiment. ``ES Test" and ``Test" are results of final model performances with and without early stopping. CG (5) is the best performing method and again IFT-Identity produces competitive second to best performance.} 
\label{tab:perparam_perfcifar}
\begin{tabular}{lllll}
\hline
  & \multicolumn{4}{c}{Accuracy}  \\ 
\cline{2-5}
Name   &  Training   & Validation & Test & ES Test\\
\hline
3 Step Neumann&100.00$\pm$0.00\%&86.00$\pm$19.80\%&18.00$\pm$3.27&18.67$\pm$6.60\\
10 Step Neumann&100.00$\pm$0.00\%&51.33$\pm$2.49\%&15.33$\pm$0.94&20.00$\pm$2.83\\
20 Step Neumann&100.00$\pm$0.00\%&35.33$\pm$17.46\%&18.00$\pm$2.83&17.33$\pm$6.60\\
35 Step Neumann&95.33$\pm$0.94\%&16.67$\pm$0.94\%&14.67$\pm$0.94&14.00$\pm$1.63\\
50 Step Neumann&77.33$\pm$0.94\%&18.00$\pm$0.00\%&8.00$\pm$0.00&8.00$\pm$0.00\\
3 Step CG&100.00$\pm$0.00\%&94.67$\pm$0.94\%&20.00$\pm$3.27&23.33$\pm$1.89\\
5 Step CG&100.00$\pm$0.00\%&92.00$\pm$0.00\%&18.67$\pm$0.94&24.67$\pm$0.94\\
10 Step CG&82.67$\pm$5.25\%&14.67$\pm$0.94\%&4.00$\pm$0.00&10.00$\pm$0.00\\
IFT-Identity&100.00$\pm$0.00\%&84.67$\pm$2.49\%&20.00$\pm$2.83&21.33$\pm$0.94\\
T1-T2&100.00$\pm$0.00\%&88.67$\pm$1.89\%&20.67$\pm$1.89&-\\
\hline
\end{tabular}
\end{table}
By looking at tables \ref{tab:perparam_perfmnist} and \ref{tab:perparam_perfcifar}, one would notice that with the Neumann (3) approximation, the MNIST validation set is successfully overfitted. However, on the CIFAR-10 dataset, the same approximation would seem to not overfit. Nevertheless, CIFAR-10 results are over 3 repetitions of the experiment. Two out of the three experiments successfully overfitted the dataset except for a single experiment which on the validation set reached a maximum validation of 58\%. IFT-Identity and T1-T2 performances are surprisingly and consistently close to that of the best performing approximation - Neumann (3).\\
Increasing the number of Neumann terms paints the algorithm more unstable and exacerbates the ability of the model to overfit the validation set. When the average validation accuracy is high, as in Neumann (50) for MNIST, one would find that the variance is extremely high for an acceptable margin in Deep Learning training. Similarly, increasing the number of CG steps worsens its' performance. This counter-intuitive observation will be studied and explained in section \ref{sec:approxacc} with relation to the method's ability in approximating the loss function curvature after convergence.\\
\textbf{Generalization Power:} As previously stated, if the training procedure is trained further after overfitting the validation set, a catastrophic forgetting occurs; both validation loss and accuracy start worsening. Therefore, all the above experiments were repeated with early stopping conditioned on the validation loss not improving for 15 meta-updates. All method's generalization power improves if we condition the training on the validation loss except for CG (5)\&(10) on MNIST and Neumann (20)\&(35) on CIFAR-10. This further worsening of performance is because those particular methods either exhibit very high variance regarding how long the method takes before converging or are unstable. The first issue means that the method could stuck in a local optimum for a while before escaping again and early stopping is placed to terminate too soon. However, this is a sign of the approximation method being unrobust. Robustness of training convergence speed and stability of approximating inverse Hessian-vector products are discussed and investigated further in sections \ref{sec:phase1cost} and \ref{sec:approxacc} respectively. One can safely conclude that it would be a beneficial practice to \textit{\textbf{place early stopping on validation loss when learning some meta-knowledge using Implicit Meta-Learning}}.\\
The best performing methods are Neumann (3) and CG (5) on MNIST and CIFAR-10 respectively. Solutions produced by the IFT-Identity approximation are second best. Nevertheless, they overall have equal or lower variance than the best solution found. Hence IFT-Identity has consistently produced stable and consistent performance across both datasets, despite not over-fitting the validation data. T1-T2 generalization is unclear (better than Neumann(3) on CIFAR-10 and worse on MNIST). This will be investigated further in section \ref{sec:perlayer} experiments.
\subsection{Approximation Methods Comparison to Exact Inverse Hessian-Vector Product}\label{sec:approxacc}
\begin{figure}[tb]
\begin{center}
\centerline{\includegraphics[width=1\columnwidth]{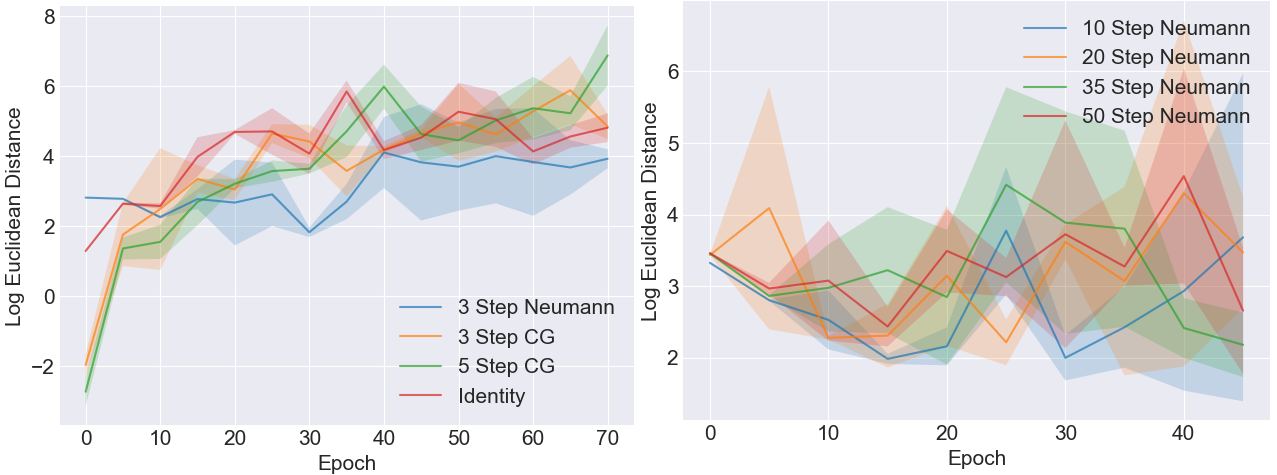}}
\caption{Comparison of approximations methods with respect to exact inverse Hessian-vector product.}
\label{fig:hessiancomparison}
\end{center}
\end{figure}

Computing the Hessian and its exact inverse was impossible for a simple MLP network on the MNIST dataset. It requires allocating \textbf{1.55 Terabytes} for an MLP with 784 input and hidden neurons. Therefore, a smaller MLP with 32 hidden neurons was used to enable computing the exact Hessian and its inverse. Furthermore, using the full resolution of MNIST made computing the inverse unstable, therefore the images were resized down to 14$\times$14 pixels. Using the convention of reporting the highest validation loss achieved, summary performance statistics are shown in table \ref{tab:mnistperparamresized}. The validation losses are healthier and greater than expected for the small capacity of the MLP employed here. Therefore, conclusions made here can be extrapolated to the full setting with larger MLPs. Finally, using the same setup of learning per MLP weight $L_2$ regularizer on MNIST, the exact inverse Hessian-vector product was computed every 5 meta-updates giving 20 comparisons in total. Computing the exact inverse of the Hessian was sometimes unstable throwing CUDA overflow exceptions among others. Hence, unstable computations were discarded. Comparisons are shown in figure \ref{fig:ablation1}\\
\begin{table}
\centering
\caption{\textbf{MNIST} - Per Weight Regularizer with Small MLP Experiment (reporting mean and std values). ``Best Step @" is a synonym to Convergence Epoch from previous tables. Computing the inverse of the exact Hessian was occasionally unstable. Hence, the absence of its results.} 
\label{tab:mnistperparamresized}
\begin{tabular}{lllll}
\hline
 & \multicolumn{3}{c}{Accuracy}  \\ 
\cline{2-4}
Name   &  Training & Validation & Test & Best Step @\\    \hline
3 Step Neumann&43.33$\pm$1.89\%&42.00$\pm$6.53\%&19.33$\pm$2.49&15.00$\pm$5.10\\
10 Step Neumann&44.00$\pm$3.27\%&46.67$\pm$4.99\%&25.33$\pm$8.38&19.33$\pm$7.41\\
20 Step Neumann&51.33$\pm$6.18\%&43.33$\pm$4.99\%&27.33$\pm$12.26&40.33$\pm$26.71\\
35 Step Neumann&41.33$\pm$3.40\%&38.00$\pm$4.90\%&18.00$\pm$2.83&18.00$\pm$5.35\\
50 Step Neumann&45.33$\pm$6.18\%&38.00$\pm$6.53\%&31.33$\pm$12.26&40.33$\pm$40.37\\
3 Step CG&54.00$\pm$6.53\%&52.67$\pm$6.60\%&34.00$\pm$7.12&69.00$\pm$36.25\\
5 Step CG&58.67$\pm$0.94\%&47.33$\pm$2.49\%&40.00$\pm$0.00&54.33$\pm$33.51\\
Identity&52.00$\pm$2.83\%&42.67$\pm$6.60\%&31.33$\pm$5.25&38.00$\pm$40.32\\
\hline
\end{tabular}
\end{table}
\textbf{Observation 1:} \textit{Increasing Neumann Series terms worsens its accuracy:}\\
To understand why, please note that Neumann Series is an iterative shrinking of the direct validation loss gradient (algorithm \ref{alg:imlift}). By assumption, the base model was trained until convergence and hence its validation loss gradient is already extremely small. Therefore, Neumann Series introduces sparse matrix multiplication in the hypergradient calculation, and increasing Neumann terms quickly introduce vanishing gradients. Moreover, the shrinking is administered by subtracting a weighted and exponentiated Hessian-vector product where the vector multiplied by is the direct validation loss. The weight, $\alpha$ is set to a default of 0.1. Therefore, these multiple exponentiations and weighting of matrices with low norms quickly introduce floating-point errors among other numerical stabilities. These factors altogether produced the phenomenon evident in figure \ref{fig:hessiancomparison}.\\
\textbf{Observation 2:} \textit{Catastrophic forgetting observed when IML is trained beyond convergence can be explained in terms of the accuracy of estimating inverse Hessian-vector products:}\\
Neumann(3) only exhibits high variance after the algorithm convergences. Few Neumann terms provide the most consistent and robust estimate. This variance in estimating the inverse Hessian-Vector product is one source of explanation to why training and validation performance worsens and the need for early stopping.
\subsection{Compute Cost}\label{sec:phase1cost}
\begin{wraptable}{r}{6.5cm}
\caption{Maximum memory allocated for the function calls computing the inverse Hessian-vector product for both MNIST and CIFAR-10 datasets in Mega, Giga, and Terabytes (Mb, Gb, Tb). Both Neumann Series and CG allocate constant memory irrespective of the number of iterations/terms. Row ``Exact" denotes the exact Hessian computation.}\label{tab:perparam_memory}
\begin{tabular}{ccc}
\hline
Name &  Data  &  Memory\\
\hline
Neumann & MNIST & 58.35 Mb\\
Neumann & CIFAR-10 & 875 Mb\\
CG & MNIST & 73.18 Mb\\
CG & CIFAR-10 & 1101 Mb\\
Identity & MNIST & 0.0\\
Identity & CIFAR-10 & 0.0\\
Exact& MNIST& 1554 Gb\\
Exact& CIFAR-10& 358 Tb\\
\hline
\end{tabular}
\end{wraptable}
\textbf{Memory Consumption:} All methods consume a constant pre-allocated memory for approximation functions calls. Memory costs are shown in table \ref{tab:perparam_memory}. Since multiplying by Identity yields the same matrix, then the inverse Hessian-vector product computation is skipped and no extra memory is allocated for it. Hence, the memory cost for the Identity approximation is zero. Regarding table \ref{tab:perparam_memory}, note that memory costs difference between Neumann and CG unclear on the MNIST dataset materialize on CIFAR-10. This is because the CIFAR-10 images have 3 channels each with more pixels than the one channeled MNIST images.\\
\textbf{Compute time} needed for the experiments on both both MNIST and CIFAR-10 are shown in table \ref{tab:perparamtimemnist} and \ref{tab:perparamtimecifar} respectively. For T1-T2, measurements are computed over 50 steps to normalize comparison metrics. The epoch column is the time taken for the whole meta-epoch to run, including the inner loop steps, full hypergradient computation, and meta-update. The Approximation column is the time taken for approximating the inverse Hessian-vector product. Finally, the Convergence Epoch is the total number of epochs taken to achieve the best validation error as reported previously.\\
\begin{table}
\centering
\caption{\textbf{MNIST:} Per Weight Regularizer Experiment Time Cost. Reported values are mean and standard deviations measured in milliseconds for ``time cost".} 
\label{tab:perparamtimemnist}
\begin{tabular}{llll}
\hline
  & \multicolumn{2}{c}{Time Cost}  \\ 
\cline{2-3}
Name   &  Epoch & Approximation & Convergence Epoch\\
\hline
3 Step Neumann&186.34$\pm$4.68 ms&5.18$\pm$0.04 ms&133.33$\pm$47.84\\
10 Step Neumann&191.01$\pm$1.43 ms&16.45$\pm$0.07 ms&30.67$\pm$2.62\\
20 Step Neumann&200.90$\pm$1.30 ms&31.96$\pm$0.20 ms&24.67$\pm$1.25\\
35 Step Neumann&226.15$\pm$0.29 ms&56.31$\pm$0.10 ms&35.33$\pm$8.34\\
50 Step Neumann&248.78$\pm$1.59 ms&80.03$\pm$0.56 ms&31.33$\pm$11.26\\
3 Step CG&164.26$\pm$1.17 ms&16.60$\pm$0.06 ms&155.33$\pm$133.41\\
5 Step CG&169.49$\pm$0.46 ms&25.94$\pm$0.07 ms&354.33$\pm$168.69\\
10 Step CG&193.15$\pm$0.46 ms&49.86$\pm$0.10 ms&0.00$\pm$0.00\\
Identity&175.03$\pm$1.14 ms&0.20$\pm$0.01 ms&146.33$\pm$77.81\\
T1-T2&464.00$\pm$6.5 ms&0.22$\pm$0.00 ms&409.00$\pm$215.99\\
Exact Hessian& - & 120.15$\pm$211.83 s &-\\
\hline
\end{tabular}
\end{table}
\begin{table}
\centering
\caption{\textbf{CIFAR-10:} Per Weight Regularizer Experiment Time Cost. Reported values are mean and standard deviations measured in milliseconds for ``time cost".} 
\label{tab:perparamtimecifar}
\begin{tabular}{llll}
\hline
  & \multicolumn{2}{c}{Time Cost}  \\ 
\cline{2-3}
Name   &  Epoch & Approximation & Convergence Epoch\\
\hline
3 Step Neumann&797.01$\pm$0.21 ms&22.98$\pm$0.01 ms&106.33$\pm$47.59\\
10 Step Neumann&850.93$\pm$0.06 ms&76.67$\pm$0.01 ms&17.33$\pm$2.49\\
20 Step Neumann&927.41$\pm$0.83 ms&153.37$\pm$0.05 ms&113.33$\pm$106.15\\
35 Step Neumann&1043.40$\pm$0.15 ms&268.60$\pm$0.05 ms&2.67$\pm$2.05\\
50 Step Neumann&1164.73$\pm$0.54 ms&382.50$\pm$0.19 ms&2.00$\pm$0.00\\
3 Step CG&870.17$\pm$0.10 ms&97.23$\pm$0.11 ms&118.00$\pm$24.10\\
5 Step CG&929.13$\pm$0.20 ms&156.83$\pm$0.24 ms&159.33$\pm$15.84\\
10 Step CG&1089.78$\pm$0.75 ms&307.33$\pm$0.12 ms&1.33$\pm$0.47\\
Identity&774.08$\pm$0.05 ms&0.00$\pm$0.00 ms&132.00$\pm$67.15\\
T1-T2&2124.50$\pm$2.00 ms&0.00$\pm$0.00 ms&165.79.33$\pm$52.62\\
\hline
\end{tabular}
\end{table}
The approximation time taken as a ratio of the whole epoch time grows as we use more Neumann terms and CG steps. For high-performing methods such as Neumann (3), the approximation time is negligible; 3\% of the whole epoch time. Expectedly, T1-T2 epoch time is the most expensive method as hypergradients are computed every single step of the epoch. The highest cost of Implicit Meta-Learning is the time taken to train the base model to convergence in all stable and high-performing methods. This conclusion will be revisited in meta-knowledge (2) experiments where the full datasets are used in a standard machine learning routine that aims to improve generalization.\\
When using a similar number of iterations, CG is approximately three times slower than Neumann. Focusing on a well-performing method such as Neumann(3) for both MNIST and CIFAR-10, one would note that they converge slower than most methods but while yielding better overall performance. As expected, there exists a trade-off between the training time taken of each approximation method and the respective quality of the solution found. The more stable a method is, the more time it takes before converging to produce better solutions. Close-performing methods take an approximately similar number of epochs to converge with IFT-Identity taking the most number of epochs to converge.\\
What seems to be fast Convergence in some methods is in effect a sign of instability. For example, CG (10), which consistently produced random test performances, appears to converge early. This translates to the validation loss being highest after initialization before being degraded by training. Finally, there is a discrepancy in the inner loop cost when using the IFT-Identity approximation for the MNIST dataset. It is slower than CG (3) and CG (5) for example. As all experiments are run sequentially on a K80 GPU, this outlier is potentially due to CUDA not freeing some cores or resources properly at some point before that experiment and hence it ends up using fewer resources. 
\section{Learning Per Layer Regularizer}\label{sec:perlayer}
\subsection{Training Details}
This problem setup is studied to dig deeper into the generalization power of different solutions found by the approximation methods. The full training MNIST and CIFAR-10 datasets were used with their standard train/test split. The testing set was split by half for both the validation and test set which produces 5000 samples each. Batch size used was 128 samples per base model update. For the meta-update, the training and validation losses were computed on the full available data although it was found that using a small random sample of 500 instances for each gives a great approximation for the true losses.\\
Initially, IML was found to be unstable on the CIFAR-10 dataset producing random predictions. This only happened when inner loop steps are ``not long enough". In an attempt to reduce the number of inner loop steps needed, the base model was initialized using converged parameters of pre-training using the respective dataset. With this initialization, the threshold of ``enough steps" found for the base model training reduced but varied between methods. Inspecting the base model weights after meta-updates showed that when using inner training steps less than the found ``enough" steps which vary among methods then all model parameters vanish quickly making the network weight matrices extremely sparse.\\
This problem was pinpointed to be an early underestimation of the training loss in the first meta-epoch. That is, once the model starts training using the hyperparameters, the training loss initially decreases before jumping drastically again. The thresholds established manually were exactly the steps needed to reach that jump. In summary, if the base model is updated too early, the loss of the hyperparameters initialization is underestimated, assumed well-performing and the meta-update takes a step in the wrong meta-loss direction destabilizing the base model performance.\\
\textbf{Warm-Up Training Procedure:} We propose to train the base model to convergence with the initialized hyperparameters before doing meta-updates. This enables the training loss to stabilize and converge. It was found that this warm-up training procedure stabilizes the base model training as long as the warm-up training allows the jump to occur. Furthermore, it was found that every meta-update does not take the base model new loss far away from convergence, and hence the number of inner loop steps needed for the base model to converge are half that of the warm-up convergence.\\
The experiments discussed in this section had a 1000 step warm-up training of the base model with initialized hyperparameters which brings the model to convergence. Additionally, the number of inner loop steps used was 500 which allows the base model to converge after every single meta-update. Meta-knowledge was updated 50 times except for T1-T2 where the base model and meta-updates alternates (50$\times$500 total updates each). Generalization power and compute cost are again discussed. Finally, an ablation study in section \ref{sec:ablation} investigates how varying the number of inner loop steps for training the base model influences the solution found.
\subsection{Generalization Power}
\begin{table}
\centering
\caption{\textbf{MNIST} - Per layer regularizer experiment statistics (reporting mean and standard deviations). The baseline reported here used no regularization as it was found that Grid Searching for an $L_2$ did not produce better solutions (grid search results are in appendix \ref{app:baselines}).} 
\label{tab:mnistperlayer}
\begin{tabular}{llll}
\hline
 & \multicolumn{3}{c}{Accuracy}  \\ 
\cline{2-4}
Name   &  Training & Validation & Test\\    \hline
Baseline&94.78$\pm$0.10\%&94.82$\pm$0.24\%&94.42$\pm$0.13\\
\hline
3 Step Neumann&99.41$\pm$0.03\%&98.05$\pm$0.05\%&98.14$\pm$0.10\\
10 Step Neumann&99.32$\pm$0.01\%&97.97$\pm$0.07\%&98.02$\pm$0.00\\
20 Step Neumann&99.22$\pm$0.02\%&97.89$\pm$0.03\%&97.95$\pm$0.01\\
35 Step Neumann&99.07$\pm$0.02\%&97.80$\pm$0.02\%&97.91$\pm$0.03\\
50 Step Neumann&98.84$\pm$0.01\%&97.65$\pm$0.05\%&97.80$\pm$0.06\\
3 Step CG&99.44$\pm$0.02\%&98.25$\pm$0.07\%&98.06$\pm$0.04\\
5 Step CG&99.26$\pm$0.10\%&98.15$\pm$0.01\%&98.17$\pm$0.07\\
10 Step CG&33.73$\pm$25.44\%&34.34$\pm$25.47\%&10.73$\pm$0.22\\
Identity&99.45$\pm$0.02\%&98.06$\pm$0.07\%&98.10$\pm$0.08\\
T1-T2&23.38$\pm$2.86\%&23.25$\pm$2.81\%&10.15$\pm$0.92\\
\hline
\end{tabular}
\end{table}
\begin{table}
\centering
\caption{\textbf{CIFAR-10} - Per Layer Regularizer Experiment Statistics. The baseline reported here used no regularization as it was found that Grid Searching for an $L_2$ did not produce better solutions (grid search results are in appendix \ref{app:baselines}).} 
\label{tab:cifarperlayer}
\begin{tabular}{llll}
\hline
 & \multicolumn{3}{c}{Accuracy}  \\ 
\cline{2-4}
Name   &  Training & Validation & Test\\    \hline
Baseline&58.25$\pm$0.19\%&51.72$\pm$0.13\%&50.39$\pm$0.50\\
\hline
3 Step Neumann&98.64$\pm$0.25\%&55.63$\pm$0.03\%&55.12$\pm$0.04\\
10 Step Neumann&34.69$\pm$4.54\%&35.11$\pm$4.35\%&34.13$\pm$4.74\\
20 Step Neumann&30.80$\pm$2.58\%&31.49$\pm$2.08\%&28.39$\pm$3.61\\
35 Step Neumann&28.57$\pm$0.97\%&29.77$\pm$0.74\%&23.41$\pm$0.56\\
50 Step Neumann&27.88$\pm$2.12\%&28.84$\pm$1.78\%&23.88$\pm$0.88\\
3 Step CG&98.77$\pm$0.24\%&56.40$\pm$0.29\%&54.83$\pm$0.53\\
5 Step CG&35.59$\pm$3.90\%&36.31$\pm$4.02\%&28.29$\pm$12.87\\
10 Step CG&18.50$\pm$2.15\%&18.57$\pm$2.59\%&10.16$\pm$0.00\\
Identity&98.59$\pm$0.39\%&55.69$\pm$0.18\%&55.02$\pm$0.82\\
T1-T2&99.91$\pm$0.01\%&56.41$\pm$0.22\%&55.37$\pm$0.56\%\\
\hline
\end{tabular}
\end{table}
Results are shown in tables \ref{tab:cifarperlayer} and \ref{tab:mnistperlayer}. Meta-Learning improves over baseline models generalization for both datasets when using either Neumann (3) or CG (3). Results for optimizing a single regularization value for the baseline using grid search are added in appendix \ref{app:baselines}. None of the searched for values performed any better than the baseline performance reported without regularization in tables \ref{tab:mnistperlayer} and \ref{tab:cifarperlayer} which shows how hard it is to optimize such simple hyperparameters.\\
\textbf{Observations Summary:} Neumann (3) and CG (3) improve over baselines in both MNIST and CIFAR-10 experiments. Consistent with findings of the previous section, increasing Neumann series or CG iterations degrades performance, using CG introduces higher variance than Neumann series does, and using the IFT-Identity approximation yields competitive results to best-performing methods on both experiments.\\
\textbf{Stability Issues:} Some seeds for the MNIST experiments produced unstable results across all approximations and it was hard to find working seeds. Therefore, some of these results are over two repetitions of the experiments. The exception is T1-T2 on MNIST which as shown in table \ref{tab:mnistperlayer} is unstable producing random predictions across all seeds. The observed algorithm instability only occurred on the MNIST dataset.\\
\textbf{Conclusion:} When the meta-knowledge learned is not over-parameterized, training beyond validation loss convergence does not worsen performance. The Validation and Test performance are comparatively close which shows that choosing reasonable hyperparameters as the meta-knowledge does not overfit but rather improves performance over baseline. The improvement over baselines is significantly larger than what one would expect on both datasets.
\subsection{Compute Cost}
\textbf{Memory Cost:} In \ref{tab:perlayer_memory}, memory consumed is much higher than the overfitting a small validation set experiments although the meta-knowledge learned is smaller. This is a direct result of the batch sizes being four times larger and approximations retaining the backpropagation graphs in memory during each step. In addition, unlike the previous experiments, Neumann series is using a higher memory than CG in this setup although not significantly larger. This is a direct result of copying the accumulated validation loss gradient (algorithm \ref{alg:imlift}) which is produced for a set 500 times larger than before.\\
\begin{wraptable}{r}{6.5cm}
\caption{Maximum memory allocated for the function calls computing the inverse Hessian-vector product for both MNIST and CIFAR-10 datasets. Both Neumann Series and CG allocate constant memory irrespective of the number of iterations/terms}\label{tab:perlayer_memory}
\begin{tabular}{ccc}
\hline
Name &  Data  &  Memory\\
\hline
Neumann & MNIST & 1.6 GB\\
Neumann & CIFAR-10 & 5.6 Gb\\
CG & MNIST & 1.4 Gb\\
CG & CIFAR-10 & 5.2 Gb\\
Identity & MNIST & 0.0\\
Identity & CIFAR-10 & 0.0\\
\hline
\end{tabular}
\end{wraptable}
\textbf{Time cost} for CIFAR-10 in learning per layer $L_2$ regularization weight is shown in table \ref{tab:cifarperlayertime}. Focusing on a well-performing method such as Neumann(3), the approximation only takes 25\% of the total meta-iteration time. Therefore, as the inner loop task becomes more complex, the most costly training overhead is allowing the base model to converge. Even though the proposed warm-up procedure helps reduce the number of inner loop steps taken to converge by half, it is still costly to run both the warm-up and inner loop to convergence routines in modern models such as ViT \cite{dosovitskiy2021an}. Therefore, when repeatedly training the base model till convergence is impractical, one would resort to using fewer base model training steps. In the next section, an ablation study that simulates such a situation is conducted.
\begin{table}
\centering
\caption{\textbf{CIFAR-10} - Per Layer Regularizer Experiment Time Cost.} 
\label{tab:cifarperlayertime}
\begin{tabular}{llll}
\hline
  & \multicolumn{2}{c}{Time Cost}  \\ 
\cline{2-3}
Name   &  Epoch & Approximation & Convergence Epoch\\    \hline
3 Step Neumann&12.99$\pm$5.03 s&3.10$\pm$0.20 s&30.33$\pm$5.56\\
10 Step Neumann&20.25$\pm$28.86 s&10.36$\pm$3.59 s&32.00$\pm$19.20\\
20 Step Neumann&30.64$\pm$17.30 s&20.75$\pm$8.72 s&5.33$\pm$2.87\\
35 Step Neumann&46.20$\pm$30.45 s&36.29$\pm$49.78 s&16.67$\pm$18.62\\
50 Step Neumann&61.80$\pm$128.05 s&51.88$\pm$108.90 s&4.00$\pm$0.82\\
3 Step CG&18.79$\pm$19.24 s&9.65$\pm$6.26 s&36.33$\pm$6.55\\
5 Step CG&23.99$\pm$812.48 s&14.91$\pm$742.76 s&32.33$\pm$21.48\\
10 Step CG&35.78$\pm$32.13 s&26.84$\pm$29.19 s&0.33$\pm$0.47\\
Identity&9.86$\pm$67.73 s&0.00$\pm$0.00 s&41.67$\pm$5.25\\
T1-T2&55$\pm$75 s&0.00$\pm$0.00 s&30.426$\pm$10.41\\
\hline
\end{tabular}
\end{table}
\subsection{Ablation Study}\label{sec:ablation}
\begin{figure}[tb]
\begin{center}
\centerline{\includegraphics[width=1\columnwidth]{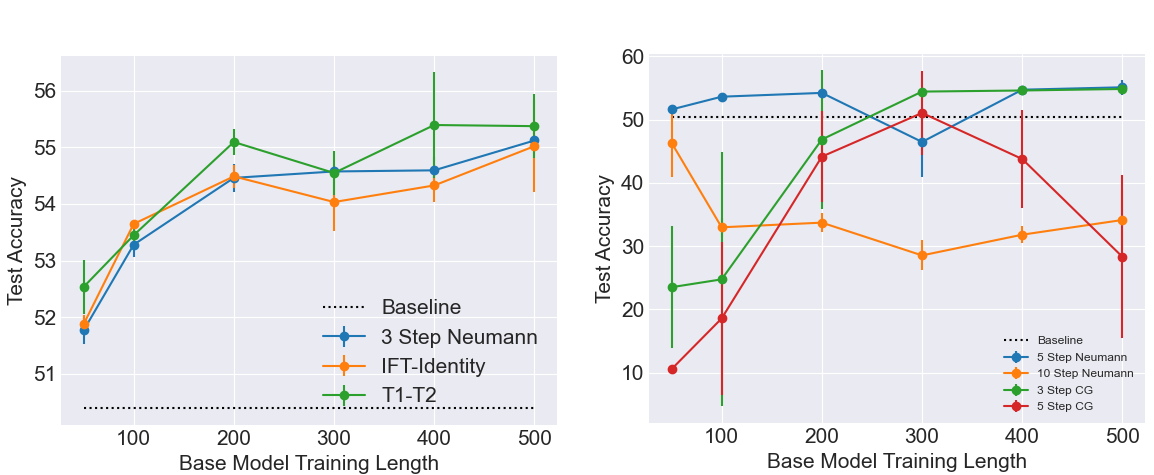}}
\caption{Generalization power of IML for Learning Per Layer $L_2$ Regularizer weight on the CIFAR-10 Dataset when using different inverse Hessian-vector product approximation methods while varying number of base model training steps.}
\label{fig:ablation1}
\end{center}
\end{figure}

\textbf{Varying Inner Steps:} Same warm-up procedure was run as explained earlier to eliminate the instability associated with initialization and control comparison. The training steps of the base model after each meta-update were varied from 50 to 500 steps (starting with 50, 100 and then up to 500 with 100 step increments). To simulate the situation where compute resources are scarce, the number of meta-updates are kept constant (50 total updates) while decreasing base model training steps. Comparing accuracies across models trained for different number of total steps is conducted to directly assess how much accuracy is sacrificed in such situations where computational resources are scrace. In the case of T1-T2, the question becomes if reducing total number of training steps while meta-updating after each inner loop step is necessary in comparison to updating less frequently as in IFT-Identity. Results are shown in figure \ref{fig:ablation1}.\\
First, starting from 100 inner loop steps with 50 meta-updates, the validation loss, and hence the outer loop, converges for all methods (plots included in appendix \ref{app:ablation_evolution_plots}). Therefore, learning stability of the meta-knowledge is not sensitive to the number of inner loop steps. Using fewer steps only affects the quality of the solution found. Note that all methods shown in the right plot of figure \ref{fig:ablation1} are not consistently improving over the baseline. Meanwhile, methods in the left plot are always outperforming the baseline. Concerning the left plot, T1-T2 is occasionally performing better than IFT-Identity and Neumann (3) but not consistently. To make the most of using T1-T2, one has to figure out the optimal steps needed to train the base model for which is costly and challenging. As shown in table \ref{tab:cifarperlayertime}, the compute cost of each 500 T1-T2 steps is at least 4x that of Neumann (3) and at least 5x of IFT-Identity.\\
With the noise present in the base model losses estimation, T1-T2 explores the meta-loss landscape much more aggressively with unnecessary overhead. On the other hand, training the base model for longer provides a better estimation of the meta-knowledge performance which produces a better hypergradient direction allowing each meta-update to navigate the meta-loss more efficiently reaching a solution faster at the expense of slightly lower performance.
\section{Discussion}
\subsection{Summary}
A great deal of attention has been given recently to Implicit Gradients for Meta-Learning (IML) \cite{lorraine2019optimizing,rajeswaran2019metalearning,ren2020unlabeled}. Such methods require computing second-order derivatives which are impractical for modern-day Deep Learning models. Previous research resort to various approximation methods to estimate such inverse Hessian-Vector products. In this chapter, these approximation methods, namely CG, Neumann Series, and IFT-Identity, were compared and contrasted in terms of compute cost, stability, generalization power, and utilization power of the available validation data. T1-T2 \cite{pmlr-v48-luketina16} as a famous method was too investigated. First, the problem of overfitting a small validation set was used to study how different approximations utilize the validation set. Overfitting was induced by learning a separate $L_2$ regularization weight for each network parameter. Then, learning an $L_2$ regularization weight per each network layer experiment was conducted using the full datasets to investigate how or if IML improves over baseline generalization.\\
We have shown that when trying to overfit a small validation set, IML succeeds to do so with Neumann (3). Nevertheless, catastrophic forgetting phenomenon is always produced by IFT updates if trained beyond convergence. This was shown to be directly linked to the inability of IFT to approximate the curvature of the loss landscape accurately. The Hessian approximations beyond convergence had high variances leading the meta-update to move away from the space of the solution found. This observation is particularly of interest to the continual learning research \cite{antoniou2021defining}. In addition, we have shown how using more terms or iterations for Neumann Series or CG produces worse solutions and destabilizes training.\\
Generalization-wise, IFT-Identity was consistently producing performances at par with the best performing method in any experiment. Gain in performance for T1-T2 in comparison with less frequent updating methods such as IFT-Identity seems insignificant when the cost of the updates is reckoned with. Finally, the memory cost of all methods grows with respect to the base model batch sizes, and hence size of accumulated gradients, rather than with the size of the meta-knowledge itself. Moreover, in such experiments, training the base model to convergence becomes the highest computational cost of the experiment. Warm-up training allows reducing the inner loop number of steps needed to converge. Using fewer inner loop steps and breaking the convergence assumption still produces results that improve over the baselines without affecting meta-knowledge training stability and convergence.
\subsection{Critique and Future Work}
The approximation methods' influence on shaping the loss landscape has not been studied in this dissertation. Recently, the influence of various network architecture elements and training procedures on deep learning dynamics was studied through the lens of the evolution of the Hessian during training \cite{pmlr-v97-ghorbani19b,10.5555/3327345.3327403,alain2018negative}. With the release of recent libraries providing fast computations of the eigenvalue decomposition of deep network Hessians \cite{yao2020pyhessian}, one can better examine the sources of catastrophic forgetting observed in IML, investigate the degrading estimation ability of different approximation methods as training proceeds and question how the approximations shape and get shaped back by the loss landscape. Insights gained from such an exploration can be used to design better approximation methods, potentially produce a new method altogether or help design a new regularization scheme for the meta-loss itself.

%% file: 5.MetaFixMatch.tex
\chapter{Confidence Network for FixMatch}\label{chapter:fixingfixmatch}
\section{Introduction}
Recent best performing Semi-Supervised Learning (SSL) algorithms \cite{mpl,kim2021selfmatch} place equal importance on all available unlabelled datapoints. It was found, however, that the performance of SSL algorithms substantially worsen when the unlabelled dataset contains out-of-distribution or out-of-domain samples \cite{oliver2019realistic}. Therefore, recent attempts to address this problem has focused on learning a per instance transductive weight that reflects the importance of such sample for improving the SSL algorithm's performance \cite{ren2020unlabeled}.\\
We propose here a novel SSL algorithm that learns to inductively weigh unlabelled instances by learning a direct representation which maps images to importance/relevance weights. In a meta-learning setting, we initiate and train a Confidence Network, $\mathcal{C}(u;\omega)$, parameterized by $\omega$, whose output resembles how confident is the CN that the unlabelled image, $u$, would improve base classifier performance. We conjecture that such setup would allow the CN to extract domain-centric features from unlabelled images, $u$, prune out-of-distribution data and further use the representation to transfer knowledge to other downstream tasks. Preliminary results show that our original proposed method improves FixMatch baseline performance and learns to discerns unlabelled images relevance to the domain being studied.\\
The problem formulation, learning rules and algorithm are introduced in section \ref{sec:fixmatch_algorithm}. Experiment setup, training details and results are covered in section \ref{sec:fixmatch_results}. Subsequently, the learnt representation is interrogated and investigated in section \ref{sec:fixmatch_interrogate} to answer the question: ``What did the Confidence Network really learn?". Finally, critique of the current proposal and directions for further work are covered in section \ref{sec:fixmatch_future}.

\section{Algorithm}\label{sec:fixmatch_algorithm}
For semi-supervised setting with meta-learning, available data is divided into three sets; a fully labelled training data $\mathcal{D}=\{(x,y)\}$, an unlabelled set $\mathcal{U}=\{u\}$ and a validation set $\mathcal{V}=\{(x,y)\}$. The base classifier is a deep network $f_\theta(x)$, parameterized by $\theta$ and the Confidence Network learns to weigh unlabelled datapoints to improve base classifier performance. More formally, the following bi-level optimization problem is addressed (base model parameters dependence on CN parameters is denoted $\theta^*(\omega)$):
\begin{equation}
    \omega^*=\argmin_\omega\mathcal{L}_s(\mathcal{V}, \theta^*(\omega))=\frac{1}{|\mathcal{V}|}\sum_{i=1}^{|\mathcal{V}|}L_s^{(i)}(\theta^*(\omega))
\end{equation}

\begin{dmath}\label{eq:metafix_inner_loss}
    \theta^*(\omega)=\argmin_\theta\mathcal{L}^{train}(\mathcal{D}, \mathcal{U}, \theta,\omega)\\
    =\argmin_\theta\mathcal{L}_s(\mathcal{D};\theta)+\sum_{u\in\mathcal{U}}\mathcal{C}(\alpha(u);\omega)\mathcal{L}_u(u;\theta)
\end{dmath}
Please recall that FixMatch unsupervised loss was defined as $\mathcal{L}_u(u;\theta)=\mathbbm{1}(max(q)\geq\tau)\text{H}(\hat{q}, f_\theta(y|\mathcal{A}(u)))$ where $\alpha(u)$ and $\mathcal{A}(u)$ are FixMatch weak and strong augmentation respectively on images $u$, $q=f_\theta(y|\alpha(u))$, $\hat{q}=argmax(q)$, and $\mathbbm{1}(max(q)\geq\tau)$ filters out poor pseudo-labels. Weak augmentations include translation and rotation while strong ones are RandAugment \cite{NEURIPS2020_d85b63ef}.\\
\textbf{Update Rules: } the base model and confidence network parameters update rules at step $t$ are:
\begin{equation}\label{eq:ift_mwn_w_update}
    \omega^{(t+1)}=\omega^{(t+1)}-\beta\nabla_\omega\mathcal{L}_s(\mathcal{V}; \theta(\omega^*))
\end{equation}
\begin{equation}\label{eq:ift_mwn_theta_update}
    \theta^{(t+1)}=\theta^{(t)}-\frac{\alpha}{n}\left[\nabla_{\theta^{(t)}}\mathcal{L}_s(\mathcal{D}; \theta^{(t)})+\sum_{u\in\mathcal{U}}\mathcal{C}(\alpha(u);\omega)\nabla_{\theta^{(t)}}\mathcal{L}_u(u;\theta)\right]
\end{equation}
\textbf{Meta-Knowledge Update Rule: }Using Implicit Meta-Learning from \textit{Lorraine et al.} \cite{lorraine2019optimizing}, the Confidence Network parameters updates rules can be derived as:
\begin{dmath}
    \frac{\partial\mathcal{L}^*_s(\mathcal{V},\theta^*(\omega))}{\partial\omega}=\frac{\mathcal{L}_s(\mathcal{V},\theta^*(\omega))}{\partial\omega}+\frac{\partial\mathcal{L}_s(\mathcal{V},\theta^*(\omega))}{\partial\theta^*(\omega)}\frac{\partial\theta^*(\omega)}{\partial\omega}
\end{dmath}
where
\begin{equation}
    \frac{\partial\theta^*(\omega)}{\partial\omega}=-\left[\frac{\partial^2\mathcal{L}^{train}}{\partial\theta\partial\theta^T}\right]^{-1}\times\frac{\partial\mathcal{L}^{train}}{\partial\theta\partial\omega^T}\biggr\rvert_{\omega',\theta^*(\omega')}
\end{equation}
Therefore, equation \ref{eq:ift_mwn_w_update} becomes:
\begin{dmath}\label{eq:ift_w_final_update}
    \omega^{(t+1)}=\omega^{(t+1)}+\beta\left[
    \frac{\partial\mathcal{L}_s(\mathcal{V},\theta^*(\omega))}{\partial\theta^*(\omega)}\times
    \left[\frac{\partial^2\mathcal{L}^{train}}{\partial\theta\partial\theta^T}\right]^{-1}\times\frac{\partial\mathcal{L}^{train}}{\partial\theta\partial\omega^T}\biggr\rvert_{\omega',\theta^*(\omega')}
    \right]\\
    =\Theta^{(t+1)}+\beta\left[
    \frac{\partial\mathcal{L}_s(\mathcal{V},\theta^*(\omega))}{\partial\theta^*(\omega)}\times
    \left[\frac{\partial^2\mathcal{L}^{train}}{\partial\theta\partial\theta^T}\right]^{-1}\times\frac{\partial\mathcal{C}(\alpha(u),\omega)\mathcal{L}_u}{\partial\theta\partial\omega^T}\biggr\rvert_{\omega',\theta^*(\omega')}
    \right]
\end{dmath}
\textbf{Meta-Update Rule Interpretation:} Equation \ref{eq:ift_w_final_update} shows how the update rule orients the meta-parameters towards the ascent direction of the validation loss gradient. This is further weighted by the unsupervised loss gradient which reflects how increasing an unlabelled datapoint confidence in the batch affects that validation loss. The full training routine is shown in algorithm \ref{alg:fixing_fixmatch}.

\IncMargin{1.5em}
\begin{algorithm}[H]\label{alg:fixing_fixmatch}
\SetAlgoLined
\SetKwInOut{Input}{Input}
\SetKwInOut{Output}{Output}
\Input{Training Data $\mathcal{D}$, Validation Data $\mathcal{V}$, Unlabelled Data $\mathcal{U}$, batch sizes $n$, $p$, $m$=$|\mathcal{V}|$ and max iterations $T$.}
\Output{Classifier Network Parameters: $\theta^*(\omega^*)$ and Weighting Network parameters: $\omega^*$}
Initialize Classifier and Confidence Networks parameters as $\omega^{(0)}$ and $\Theta^{(0)}$ respectively

\While{\text{Not Converged}}{
    \For{t = 0...T-1}{
        $\{x,y\}\leftarrow$\text{SampleMiniBatch}($\mathcal{D}$, n)\\
        $\{x^{\mathcal{U}}\}\leftarrow$\text{SampleMiniBatch}($\mathcal{U}$, p)\\
        \text{Update classifier network parameters using eq \ref{eq:ift_mwn_w_update}}
    }
    $v_1=\frac{\partial\mathcal{L}_s(\mathcal{V}, \omega^*(\Theta^{(t)}))}{\omega^*(\Theta^{(t)})}$ using the validation dataset
    
    Approximate inverse Hessian-vector product, $v_2=v_1\times\left[\frac{\partial^2\mathcal{L}^{train}}{\partial\theta\partial\theta^T}\right]^{-1}$
    
    $v_3=v_2\times\frac{\partial\mathcal{C}(u;\omega)\mathcal{L}_u(u;\theta)}{\partial\theta\partial\omega}\rvert_{\omega',\theta^*(\omega')}$

    $\frac{\partial\mathcal{L}^*_s(\mathcal{V},\omega^*(\Theta))}{\partial\Theta}$=-$v_3$
    
    Compute $\Theta^{(t+1)}$ using update rule, equation \ref{eq:ift_mwn_theta_update}
    
}
 \caption{Fixing FixMatch by Confidence Network Weighting Algorithm}
\end{algorithm}

\section{Evaluation and Results}\label{sec:fixmatch_results}
\subsection{Training Details}
\textbf{Experiment Setup:} the proposed method is evaluated on the CIFAR-10 dataset. We use 250 labelled training images (25 instances per class), a validation set of size 1000 (100 instances per class) and the full training data without the validation set as unlabelled data. A relatively large validation set is being used as per the convention of \textit{Ren et al.} paper \cite{ren2020unlabeled} which places an extra annotation overhead in standard SSL. In the Critique and Further work section, a solution is proposed to remedy this concern.\\
\textbf{Evaluation Study Control:} the base model trained with CN would have implicit knowledge of extra labelled datapoints granting an unfair comparison to the baseline. Therefore, the CN was first trained, parameters producing best validation loss were saved and the trained base model was discarded. Then, the same 250 labelled training instances were used to train two FixMatch models from scratch with and without the saved CN. This strategy enables discerning the CN influence on FixMatch by comparing two models trained using the same data.\\
\textbf{Training Details:} With reference to line 1 in algorithm \ref{alg:fixing_fixmatch}, the Classifier and Confidence Networks parameters are initialized randomly. Then, a warm up routine is run to bring the Confidence Network initialization to equally weigh all unlabelled datapoints. Using a target of 1 for all unlabelled images, the Confidence Network is trained for a fixed number of steps to extract heteregeneous features from the full available data that would result in all Confidence Network predictions to be close to 1.\\
Confidence and Classifier (base model) network architecture employed was ResNet-28 \cite{zagoruyko2017wide}. Both networks were trained using SGD with momentum and weight decay of $5\times10^{-5}$. The Classification and Confidence Networks learning rates were $0.03$ and $1\times10^{-3}$ respectively. For CN, higher learning rates were also tried but were found to produce a slightly worse accuracy on the validation set.\\
FixMatch exhibits the same dynamics observed in the learning per layer $L_2$ regularization weight where the unsupervised loss increases initially before decreasing again. Therefore, the base model was trained for 210 epochs in total where the first 10 epochs were a warm-up phase allowing the unsupervised loss to stabilize. This is followed by a meta-update. Subsequently, the base model is trained for 200 epochs, updating the Confidence Network after each 5 epochs. Updating the Confidence Network after each epoch produced worse performance than updating less frequently. FixMatch also exhibits very high noise per epoch, therefore each epoch is 512 steps. In summary, after warm-up and its respective meta-update, the base model is trained for a total of 102400 steps, updating the meta-knowledge after each 2560 steps.\\
Approximating the inverse-Hessian vector product was also challenging. The Neumann Series was found to be extremely unstable when trained with and without warm-up epochs. The Confidence Network starts producing weights up to a magnitude of 200 and oscillating aggressively down to negative values and up again. Therefore, IFT-Identity approximation was resorted to. T1-T2 was attempted with and without warm-up steps, updating after each inner loop step but was too found to be unstable; the CN predicted values oscillated aggressively between 120 and -10 in consecutive steps.\\
Finally, as per equation \ref{eq:metafix_inner_loss}, the Confidence Network consumes weakly augmented images during training and original images for either the validation set or during inference when parameters are fixed. Using the weakly augmented images during training introduces variability of exposure during learning as the image augmentations are applied with probabilities. This exposes the CN to knowledge on how the augmentation is directly influencing the overall training loss.\\
\subsection{Results}
\textit{Note: CN takes approximately 16-18 hours to train for 210 epochs on a Tesla A100 Tensor-Core GPU. Given short timescale of the project, it was only feasible to provide three repetitions of the experiments as reported here for our method and the baseline.}\\
The results in table \ref{tab:phase2preliminary} show how our method improves over the baseline FixMatch performance. Please note that on experiment cifar10@250\_3 which uses random number generator seed=3, proposed method did not perform better than the baseline. It was found that the validation loss of the CN of that experiment, although monotonically decreasing, was still significantly lower than other seeds. Therefore, training for longer could have potentially improved over the baseline or a different seed would have been chosen.\\
Next questions would be what did the Confidence Network actually learn? What weights is it producing and how does those weights vary among instances? These questions are the topic of the next section. The issue that the Confidence Network requires extra labelled datapoints to train is addressed in section \ref{sec:fixmatch_future}.
\begin{table}[h]
\centering
\caption{Preliminary experimental results using only 250 labelled datapoints from the CIFAR-10 dataset. Results are reported with a Confidence Weighting Network (ours) and without (FixMatch baseline). The results shown are over three repetitions.} 
\label{tab:phase2preliminary}
\begin{tabular}{ccc}
\hline
Name   &  Baseline Test Accuracy & Ours\\    \hline
cifar10@250\_1&93.69&\textbf{94.04}\\
cifar10@250\_2&93.54&\textbf{93.71}\\
cifar10@250\_3&\textbf{93.23}&91.13\\
\hline
\end{tabular}
\end{table}

\section{What did the model learn?}\label{sec:fixmatch_interrogate}
\begin{figure}[h]
\begin{center}
\centerline{\includegraphics[width=1\columnwidth]{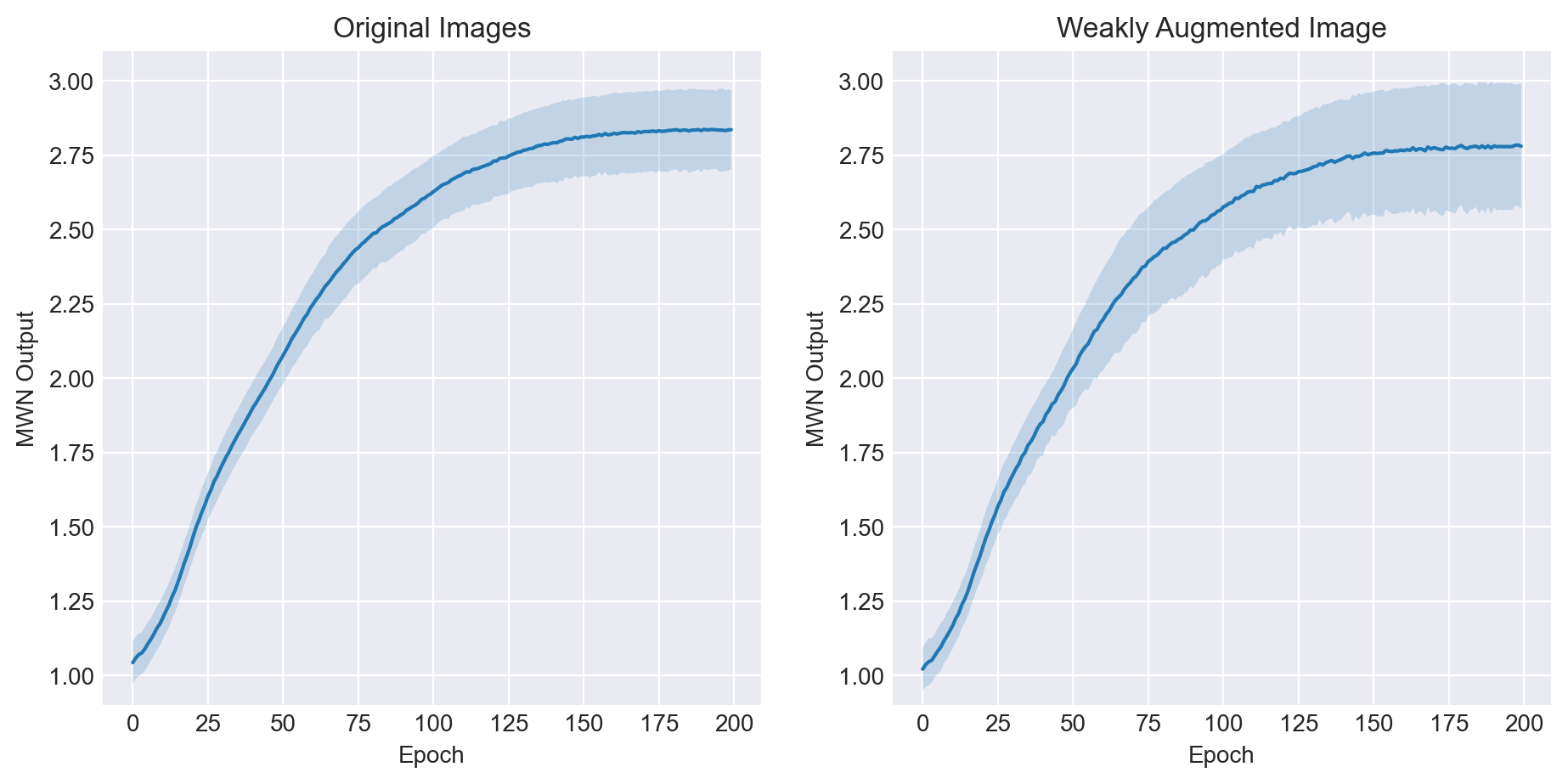}}
\caption{Average and standard deviation of weights produced by Confidence Networks for original and weakly augmented images as training proceeds.}
\label{fig:cn_outputs}
\end{center}
\end{figure}

To better understand the learnt representation of the Confidence Network, CN weights were produced for all original and weakly augmented images. All Confidence Networks produced during training were evaluated to also inspect how the outputs evolve as training proceeds. Please note that the CNs used in this section are those produced by experiment cifar10@250\_1 in table \ref{tab:phase2preliminary}.\\
The comparison is shown in figure \ref{fig:cn_outputs}. A sign for training health is the evolution of the confidence weights produced; monotonic increase before convergence. The confidence weights produced by CN for the original images is on average higher in magnitude than the Weakly Augmented. The closeness of averages can be explained in terms of the low probabilities used in the weak augmentations of FixMatch as stated earlier. Some images evaluated in the weakly augmented settings would have not been actually augmented with some probability.\\
One would expect that the weakly augmented images outputs be significantly lower. This is revealed in the high variance exhibited in the weakly augmented instances outputs. The variance of the weights reaches a magnitude as low as 2.50 which the original images output is far from. Therefore, the Confidence Network does learn to distinguish and discriminate between ``clear" and noisy instances.\\
\begin{wrapfigure}{R}{0.5\textwidth}
\begin{center}
\includegraphics[width=0.45\columnwidth]{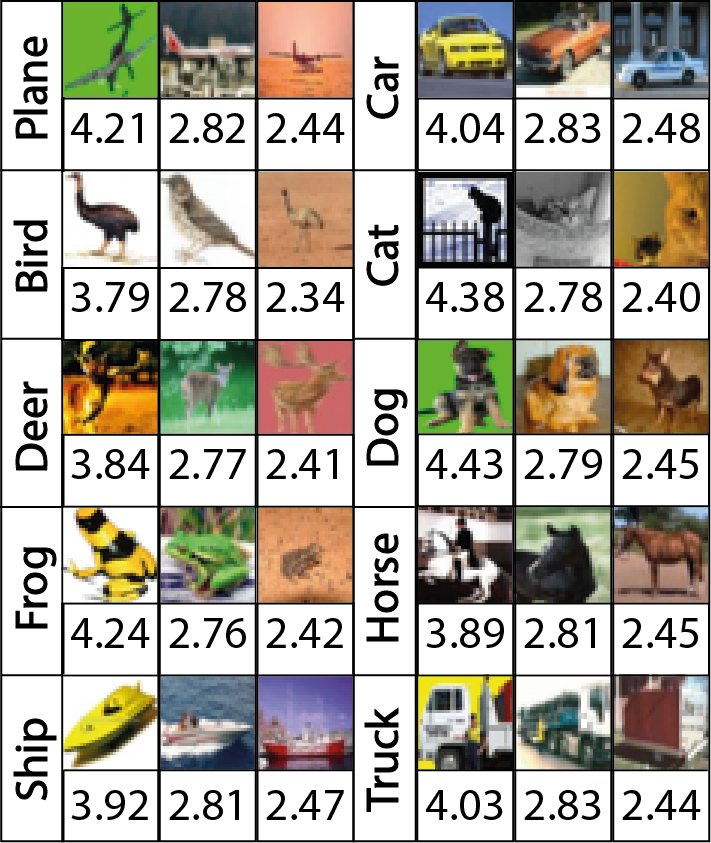}
\end{center}
\caption{From left to right per each class, the images with most, median and lowest weights predicted by Confidence Net are visualized. The number above below image is the CN predicted weight for the image}\label{fig:cn_viz}
\end{wrapfigure}

Next, we select and visualize three images per each class by the highest, median and lowest weights predicted by trained Confidence Network as per figure \ref{fig:cn_viz}. The highest and lowest weighed images are interpreted as images which the Confidence Network is most certain would increase and decrease base model performance respectively. The image with median prediction of CN is also included.\\
Note that the images with the highest inlier predictions, around the weight of 4, are all instances where the background colour is either solid or non-distracting, i.e: Truck, Frog, Airplane, Dog, Truck, Ship and Bird are all examples of this. The Confidence Network has chosen to upweigh an Automobile instance where the background is non-distracting, the car front is clear and the car colors are contrasting and downweighs instances where the angle is found to be confusing to the base network for classification. Finally, the median and weakest weights predicted are close to each other. Therefore, next we visualize the distribution of weights predicted over the whole training data.\\
\begin{wrapfigure}{R}{0.5\textwidth}
\begin{center}
\includegraphics[width=0.45\columnwidth]{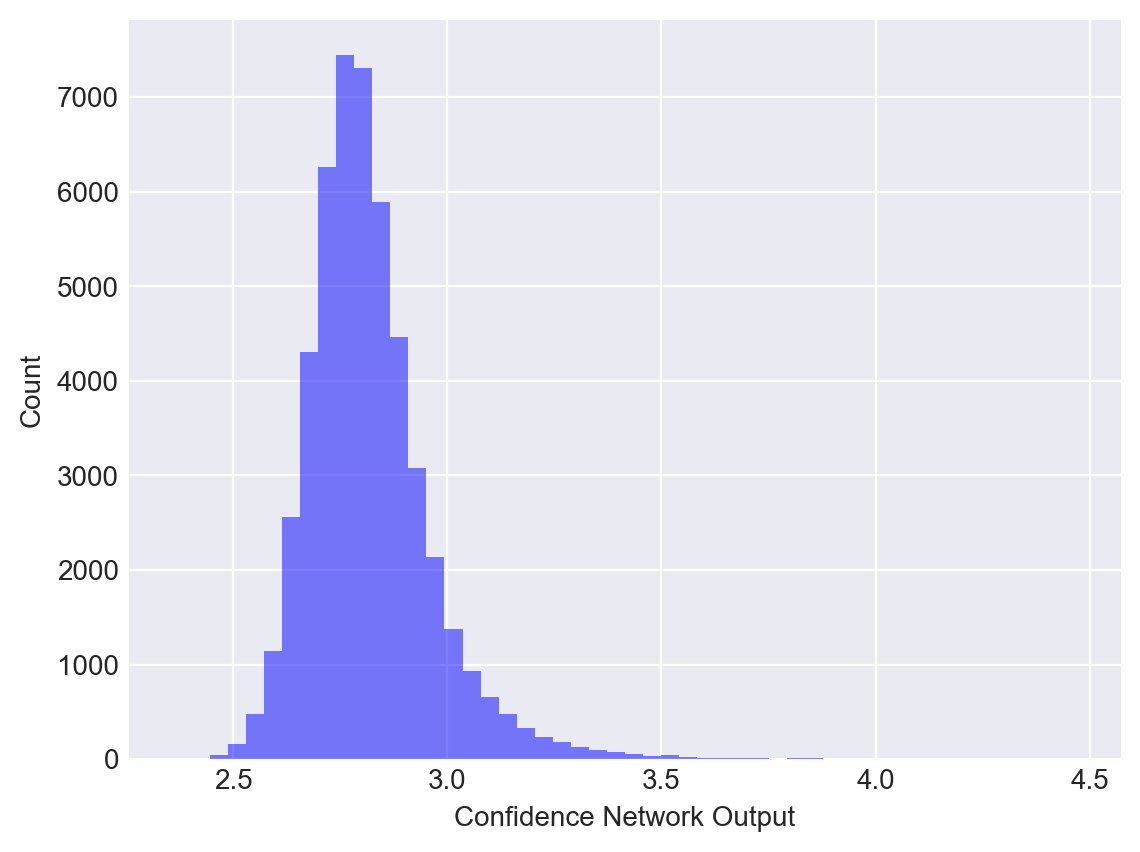}
\end{center}
\caption{Histogram over all weight predictions by the Confidence Network for the full training dataset}\label{fig:cn_hist}
\end{wrapfigure}
The histogram over all CN predictions is in figure \ref{fig:cn_hist}. Note that the CN again upweighs very few instances consistently with the observation in figure \ref{fig:cn_viz}. The average weights predicted lump around same values as in figure \ref{fig:cn_outputs}.\\
In conclusion, compelling evidence for the conjecture that the Confidence Network learns to weigh unlabelled images using domain-centric feature extraction has been provided. Some shortcomings in our proposed methodology are pinpointed in the next section. Further work to improve the algorithm is also introduced.
\newpage
\section{Conclusion and Further Work}\label{sec:fixmatch_future}
Proposed method was shown to outperform FixMatch baselines. The interpretation of the Confidence Network in section \ref{sec:fixmatch_interrogate} inspire using this formulation for other use cases such as out-of-distribution detection or domain adaptation and generalization.\\
Main shortcoming of the proposed method is the need of an extra labelled validation set. This can be addressed in two steps. First, eliminating the need for the labelled training data can be done using contrastive pre-training as proposed in SelfMatch \cite{kim2021selfmatch}. The training labelled data responsibility is to bring the base model to predict labels for the unlabelled instances with high confidence which SelfMatch solves using contrastive learning pre-training. Secondly, an ablation of the validation data size effect on base model performance must be conducted. If the Confidence Network is found to be data hungry, one potential solution would be sharing parameters between both Confidence network and base model. Then, only the final few convolutional and fully connected layers in the Confidence Network can be trained which would demand less data for the fewer trainable parameters. The final few layers of convolutional neural network are known to specialize more in domain knowledge than the initial ones. Therefore, the same interpretation instigated in this chapter can be carried forward to the new setup.\\
Finally, CIFAR-10 does not contain out-of-distribution or out-of-domain data because it was is curated with precision. Therefore, the full power of the proposed method was not conveyed or challenged. We propose to evaluate this method on a more challenging benchmark such as STL-10 which contains out-of-distribution images by design \cite{pmlr-v15-coates11a}.

%% file: 6.Conclusion.tex
\chapter{Conclusion}\label{chapter:conclusion}
\section{Summary and Reflection}
Implicit meta-learning (IML) is great in solving vanishing gradient problems and high memory costs associated with standard meta-learning. In this research, we systematically compared the largely ignored evaluation of the effect of incorporating different second order gradient approximation methods into implicit meta-learning. Approximation methods studied were Neumann series, CG and T1-T2 approximations. The methods were contrasted in terms of compute cost, generalization power, stability and power to capitalize on the available validation data.\\
We have also shown that implicit meta-learning exhibits catastrophic forgetting phenomenon if trained for few epochs after convergence. This was explained in terms of the different approximation methods inability to estimate the curvature of the loss function ($2^{nd}$ order gradient) at convergence which steers the meta-knowledge away from the solution found. Additionally, instability of IML associated with under-estimation of losses after meta-knowledge initialization was discovered and presented. A simple warm-up training procedure for initializing the base model parameters was proposed and evaluated to show that it successfully remedies the problem.\\
In the following part of the thesis, insights gained were used to design a novel semi-supervised learning algorithm which learns to weigh the consistency regularization losses of FixMatch inductively by a \textit{Confidence Network} (CN). Intuitively, CN consumes images and outputs weights reflecting how confident is the network that the instance would improve FixMatch performance. We have shown that our proposed algorithm improves over FixMatch baseline performance. Additionally, we provide compelling evidence that the CN does indeed extract domain-centric features by visualizing the distribution of it's outputs over the training data and providing example images up and down weighted. Few images with approximately twice the weight than the median output of the network were found to be ones without any distracting background information where feature extraction is much easier. 
\section{Future Work}
For the comparative analysis, we proposed to study how the different inverse Hessian-vector product approximations shape and get shaped back by the meta-loss function. An initial methodology can follow the lines of recent studies which have investigated how different architecture element and training procedures such as batch sizes affect the loss surface \cite{pmlr-v97-ghorbani19b,10.5555/3327345.3327403,alain2018negative}. There is availability of powerful tools to help such direction. For example, PyHessian provides PyTorch wrapper functions to easily and quickly compute eigenvalue decomposition, trace or highest eigenvalues of the loss function Hessian \cite{yao2020pyhessian}.\\
For our proposed semi-supervised learning algorithm, we proposed to start with an ablation study of the effect of the validation set size on algorithm performance. Thereafter, SelfMatch \cite{kim2021selfmatch} can be used in palce of FixMatch to eliminate the need of a training labelled data through contrastive pre-training. To reduce the validation set size further, sharing base model and CN parameters for the first few layers can be attempted. Finally, a dataset such as STL-10 which contains out-of-distribution samples by design should be used to produce a better evaluation of the proposed method and challenge its power.

%% file: 99.appendix_perparamplots.tex
\chapter{Overfitting a Small Validation Set}\label{app:overfitting_detailed}
\section{Detailed Experiment Results}
Naming conventions for tables in this section are as follows:
\begin{enumerate}
    \item numn\_3\_sd\_231 or CG\_3\_sd\_231 means that 3 terms/steps were used for Neumann or CG respectively and sd\_231 means that the experiment was set with seed=231 to initialize the random number generator of pytorch
    \item Per Weight (Full 1000 epochs) tables: reporting final results of the base model after the full 1000 epochs for overfitting a small validation set experiments.
    \item Per Weight (Early Stopping) tables: overfitting a small validation set with early stopping experiments results
    \item Per Weight Resized tables: results of the approximation methods comparison with respect to exact inverse Hessian-vector product experiments
    \item Per Layer tables are detailed results for the generalization experiments
    \item To differentiate this Appendix's tables from those in the body of the document, the final validation performance rather than maximum achieved is reported here
\end{enumerate}

\begingroup
\begin{table}
\tiny
\centering
\caption{\textbf{MNIST} - Per Weight (Full 1000 epochs).} 
\begin{tabular}{lllllllll}
\hline
  & \multicolumn{3}{c}{Accuracy} && \multicolumn{4}{c}{Training Cost}    \\ 
\cline{2-4} \cline{6-9} 
Exp   &  Training   & Validation & Test & & Convergence Epoch & Approximation Time &  Meta-Epoch Time T & Max Memory \\    \hline
numn\_no\_3\_sd\_231&      100.00\%&98.00\%&66.00\%&&110&5.24 ms&192.95 ms&58.35 mb\\
numn\_no\_3\_sd\_981&      100.00\%&90.00\%&60.00\%&&200&5.15 ms&183.29 ms&58.35 mb\\
numn\_no\_3\_sd\_1110&      100.00\%&98.00\%&66.00\%&&90&5.15 ms&182.77 ms&58.35 mb\\
numn\_no\_10\_sd\_231&      100.00\%&56.00\%&62.00\%&&27&16.54 ms&192.87 ms&58.35 mb\\
numn\_no\_10\_sd\_981&      100.00\%&50.00\%&44.00\%&&33&16.44 ms&190.77 ms&58.35 mb\\
numn\_no\_10\_sd\_1110&      100.00\%&60.00\%&38.00\%&&32&16.37 ms&189.39 ms&58.35 mb\\
numn\_no\_20\_sd\_231&      100.00\%&66.00\%&58.00\%&&23&31.85 ms&200.13 ms&58.35 mb\\
numn\_no\_20\_sd\_981&      100.00\%&54.00\%&42.00\%&&25&32.24 ms&202.73 ms&58.35 mb\\
numn\_no\_20\_sd\_1110&      100.00\%&66.00\%&68.00\%&&26&31.79 ms&199.84 ms&58.35 mb\\
numn\_no\_35\_sd\_231&      100.00\%&60.00\%&48.00\%&&28&56.34 ms&226.08 ms&58.35 mb\\
numn\_no\_35\_sd\_981&      100.00\%&62.00\%&58.00\%&&47&56.41 ms&226.54 ms&58.35 mb\\
numn\_no\_35\_sd\_1110&      100.00\%&54.00\%&50.00\%&&31&56.18 ms&225.83 ms&58.35 mb\\
numn\_no\_50\_sd\_231&      100.00\%&58.00\%&48.00\%&&26&79.25 ms&246.55 ms&58.35 mb\\
numn\_no\_50\_sd\_981&      12.00\%&10.00\%&10.00\%&&21&80.32 ms&249.63 ms&58.35 mb\\
numn\_no\_50\_sd\_1110&      100.00\%&56.00\%&54.00\%&&47&80.53 ms&250.16 ms&58.35 mb\\
CG\_3\_sd\_231&      100.00\%&78.00\%&56.00\%&&60&16.64 ms&165.24 ms&73.18 mb\\
CG\_3\_sd\_981&      100.00\%&76.00\%&60.00\%&&62&16.65 ms&164.91 ms&73.18 mb\\
CG\_3\_sd\_1110&      100.00\%&76.00\%&56.00\%&&344&16.52 ms&162.62 ms&73.18 mb\\
CG\_5\_sd\_231&      100.00\%&70.00\%&62.00\%&&210&26.04 ms&170.11 ms&73.15 mb\\
CG\_5\_sd\_981&      96.00\%&68.00\%&58.00\%&&591&25.86 ms&169.03 ms&73.18 mb\\
CG\_5\_sd\_1110&      100.00\%&72.00\%&66.00\%&&262&25.92 ms&169.32 ms&73.18 mb\\
CG\_10\_sd\_231&      14.00\%&16.00\%&12.00\%&&0&50.00 ms&193.80 ms&73.15 mb\\
CG\_10\_sd\_981&      14.00\%&16.00\%&12.00\%&&0&49.77 ms&192.91 ms&73.18 mb\\
CG\_10\_sd\_1110&      14.00\%&16.00\%&12.00\%&&0&49.81 ms&192.75 ms&73.18 mb\\
identity\_sd\_231&      100.00\%&82.00\%&66.00\%&&255&0.20 ms&176.09 ms&38.53 mb\\
identity\_sd\_981&      100.00\%&78.00\%&74.00\%&&77&0.19 ms&175.53 ms&38.53 mb\\
identity\_sd\_1110&      100.00\%&84.00\%&76.00\%&&107&0.21 ms&173.45 ms&38.53 mb\\
t1t2\_sd\_231&      100.00\%&90.00\%&70.00\%&&33802&0.22 ms&9.45 ms&2.71 mb\\
t1t2\_sd\_981&      100.00\%&86.00\%&66.00\%&&20196&0.22 ms&9.24 ms&2.71 mb\\
t1t2\_sd\_1110&      100.00\%&78.00\%&72.00\%&&7352&0.22 ms&9.14 ms&2.71 mb\\
\end{tabular}
\end{table}
\endgroup

\begingroup
\begin{table}
\tiny
\centering
\caption{\textbf{MNIST} - Per Weight (Early Stopping).} 
\begin{tabular}{lllllllll}
\hline
  & \multicolumn{3}{c}{Accuracy} && \multicolumn{4}{c}{Training Cost}    \\ 
\cline{2-4} \cline{6-9} 
Exp   &  Training   & Validation & Test & & Convergence Epoch & Approximation Time &  Meta-Epoch Time T & Max Memory \\    \hline
numn\_no\_3\_sd\_231&      100.00\%&100.00\%&82.00\%&&110&5.39 ms&197.44 ms&58.35 mb\\
numn\_no\_3\_sd\_981&      100.00\%&100.00\%&72.00\%&&200&5.32 ms&195.25 ms&58.35 mb\\
numn\_no\_3\_sd\_1110&      100.00\%&100.00\%&78.00\%&&90&5.33 ms&195.54 ms&58.35 mb\\
numn\_no\_10\_sd\_231&      88.00\%&68.00\%&60.00\%&&27&18.68 ms&206.96 ms&58.35 mb\\
numn\_no\_10\_sd\_981&      92.00\%&70.00\%&46.00\%&&33&17.78 ms&204.53 ms&58.35 mb\\
numn\_no\_10\_sd\_1110&      96.00\%&68.00\%&58.00\%&&32&18.19 ms&203.99 ms&58.35 mb\\
numn\_no\_20\_sd\_231&      88.00\%&68.00\%&52.00\%&&23&35.03 ms&217.03 ms&58.35 mb\\
numn\_no\_20\_sd\_981&      100.00\%&74.00\%&68.00\%&&25&33.82 ms&214.43 ms&58.35 mb\\
numn\_no\_20\_sd\_1110&      92.00\%&72.00\%&60.00\%&&26&32.78 ms&212.69 ms&58.35 mb\\
numn\_no\_35\_sd\_231&      100.00\%&74.00\%&60.00\%&&28&56.80 ms&235.50 ms&58.35 mb\\
numn\_no\_35\_sd\_981&      96.00\%&74.00\%&64.00\%&&47&55.97 ms&234.47 ms&58.35 mb\\
numn\_no\_35\_sd\_1110&      100.00\%&66.00\%&56.00\%&&31&55.53 ms&233.03 ms&58.35 mb\\
numn\_no\_50\_sd\_231&      100.00\%&68.00\%&64.00\%&&26&78.74 ms&253.42 ms&58.35 mb\\
numn\_no\_50\_sd\_981&      94.00\%&66.00\%&50.00\%&&21&78.87 ms&253.53 ms&58.35 mb\\
numn\_no\_50\_sd\_1110&      92.00\%&76.00\%&54.00\%&&47&79.14 ms&255.11 ms&58.35 mb\\
CG\_3\_sd\_231&      100.00\%&90.00\%&56.00\%&&64&16.75 ms&167.23 ms&73.18 mb\\
CG\_3\_sd\_981&      100.00\%&94.00\%&60.00\%&&50&16.73 ms&166.92 ms&73.18 mb\\
CG\_3\_sd\_1110&      100.00\%&92.00\%&56.00\%&&70&16.82 ms&167.04 ms&73.18 mb\\
CG\_5\_sd\_231&      98.00\%&54.00\%&52.00\%&&51&26.29 ms&172.95 ms&73.18 mb\\
CG\_5\_sd\_981&      98.00\%&54.00\%&54.00\%&&69&26.26 ms&172.98 ms&73.18 mb\\
CG\_5\_sd\_1110&      98.00\%&50.00\%&52.00\%&&20&26.00 ms&171.52 ms&73.18 mb\\
CG\_10\_sd\_231&      16.00\%&6.00\%&6.00\%&&0&49.98 ms&197.46 ms&73.15 mb\\
CG\_10\_sd\_981&      16.00\%&6.00\%&6.00\%&&0&50.01 ms&197.67 ms&73.18 mb\\
CG\_10\_sd\_1110&      16.00\%&6.00\%&6.00\%&&0&50.01 ms&197.26 ms&73.18 mb\\
identity\_sd\_231&      100.00\%&94.00\%&74.00\%&&85&0.23 ms&181.57 ms&38.53 mb\\
identity\_sd\_981&      100.00\%&92.00\%&78.00\%&&77&0.23 ms&180.88 ms&38.53 mb\\
identity\_sd\_1110&      100.00\%&94.00\%&78.00\%&&107&0.23 ms&181.30 ms&38.53 mb\\
t1t2\_sd\_231&      10.00\%&12.00\%&2.00\%&&12&0.21 ms&25.82 ms&2.71 mb\\
t1t2\_sd\_981&      8.00\%&10.00\%&12.00\%&&11&0.23 ms&9.51 ms&2.71 mb\\
t1t2\_sd\_1110&      16.00\%&24.00\%&20.00\%&&11&0.20 ms&9.57 ms&2.71 mb\\
\end{tabular}
\end{table}
\endgroup

\begingroup
\begin{table}
\tiny
\centering
\caption{\textbf{MNIST} - Per Weight Resized.} 
\begin{tabular}{lllllllll}
\hline
  & \multicolumn{3}{c}{Accuracy} && \multicolumn{4}{c}{Training Cost}    \\ 
\cline{2-4} \cline{6-9} 
Exp   &  Training   & Validation & Test & & Convergence Epoch & Approximation Time &  Meta-Epoch Time T & Max Memory \\    \hline
numn\_3\_sd\_231&      32.00\%&20.00\%&22.00\%&&13&4.56 ms&159.10 ms&0.75 mb\\
numn\_3\_sd\_981&      24.00\%&38.00\%&20.00\%&&10&4.47 ms&153.75 ms&0.75 mb\\
numn\_3\_sd\_1110&      28.00\%&28.00\%&16.00\%&&22&4.48 ms&153.28 ms&0.75 mb\\
numn\_10\_sd\_231&      18.00\%&16.00\%&14.00\%&&26&14.16 ms&161.72 ms&0.75 mb\\
numn\_10\_sd\_981&      42.00\%&26.00\%&34.00\%&&9&14.25 ms&163.31 ms&0.75 mb\\
numn\_10\_sd\_1110&      28.00\%&32.00\%&28.00\%&&23&14.17 ms&163.30 ms&0.75 mb\\
numn\_20\_sd\_231&      46.00\%&22.00\%&28.00\%&&19&27.80 ms&174.78 ms&0.75 mb\\
numn\_20\_sd\_981&      60.00\%&38.00\%&42.00\%&&78&27.52 ms&173.53 ms&0.75 mb\\
numn\_20\_sd\_1110&      32.00\%&24.00\%&12.00\%&&24&27.67 ms&174.20 ms&0.75 mb\\
numn\_35\_sd\_231&      18.00\%&24.00\%&14.00\%&&19&48.87 ms&197.49 ms&0.75 mb\\
numn\_35\_sd\_981&      26.00\%&30.00\%&20.00\%&&11&48.08 ms&194.49 ms&0.75 mb\\
numn\_35\_sd\_1110&      28.00\%&40.00\%&20.00\%&&24&48.56 ms&196.52 ms&0.75 mb\\
numn\_50\_sd\_231&      20.00\%&26.00\%&16.00\%&&18&69.75 ms&218.50 ms&0.75 mb\\
numn\_50\_sd\_981&      42.00\%&22.00\%&32.00\%&&6&69.52 ms&218.21 ms&0.75 mb\\
numn\_50\_sd\_1110&      54.00\%&46.00\%&46.00\%&&97&69.35 ms&217.32 ms&0.75 mb\\
CG\_3\_sd\_231&      46.00\%&38.00\%&30.00\%&&18&15.86 ms&164.27 ms&0.91 mb\\
CG\_3\_sd\_981&      60.00\%&60.00\%&44.00\%&&90&15.83 ms&163.93 ms&0.91 mb\\
CG\_3\_sd\_1110&      50.00\%&48.00\%&28.00\%&&99&15.86 ms&163.78 ms&0.91 mb\\
CG\_5\_sd\_231&      58.00\%&40.00\%&40.00\%&&10&25.07 ms&172.12 ms&0.91 mb\\
CG\_5\_sd\_981&      58.00\%&38.00\%&40.00\%&&62&24.93 ms&171.95 ms&0.91 mb\\
CG\_5\_sd\_1110&      56.00\%&50.00\%&40.00\%&&91&24.81 ms&171.23 ms&0.91 mb\\
identity\_sd\_231&      54.00\%&40.00\%&34.00\%&&8&0.21 ms&146.18 ms&0.54 mb\\
identity\_sd\_981&      46.00\%&42.00\%&36.00\%&&95&0.22 ms&142.44 ms&0.54 mb\\
identity\_sd\_1110&      50.00\%&34.00\%&24.00\%&&11&0.21 ms&143.55 ms&0.54 mb\\

\end{tabular}
\end{table}
\endgroup

\begingroup
\begin{table}
\tiny
\centering
\caption{\textbf{MNIST} - Per Layer.} 
\begin{tabular}{lllllllll}
\hline
  & \multicolumn{3}{c}{Accuracy} && \multicolumn{4}{c}{Training Cost}    \\ 
\cline{2-4} \cline{6-9} 
Exp   &  Training   & Validation & Test & & Convergence Epoch & Approximation Time &  Meta-Epoch Time T & Max Memory \\    \hline
numn\_no\_3\_sd\_231&      99.39\%&97.90\%&98.08\%&&48&157.65 ms&2230.09 ms&1600.45 mb\\
numn\_no\_3\_sd\_981&      99.37\%&97.94\%&98.06\%&&48&184.96 ms&2278.52 ms&1600.45 mb\\
numn\_no\_3\_sd\_523&      99.45\%&98.12\%&98.28\%&&49&366.54 ms&2983.20 ms&1600.45 mb\\
numn\_no\_10\_sd\_231&      99.29\%&97.88\%&98.02\%&&48&625.14 ms&2732.97 ms&1601.87 mb\\
numn\_no\_10\_sd\_981&      99.30\%&97.86\%&98.02\%&&48&638.94 ms&2827.20 ms&1601.55 mb\\
numn\_no\_10\_sd\_523&      11.24\%&10.72\%&10.88\%&&0&1266.11 ms&3863.37 ms&1600.45 mb\\
numn\_no\_20\_sd\_231&      99.18\%&97.86\%&97.96\%&&46&1235.07 ms&3288.88 ms&1601.55 mb\\
numn\_no\_20\_sd\_981&      99.22\%&97.82\%&97.94\%&&48&1235.14 ms&3295.44 ms&1601.87 mb\\
numn\_no\_20\_sd\_523&      11.24\%&10.72\%&10.88\%&&0&2549.98 ms&5115.27 ms&1601.93 mb\\
numn\_no\_35\_sd\_231&      99.00\%&97.82\%&97.88\%&&49&2162.23 ms&4238.93 ms&1601.87 mb\\
numn\_no\_35\_sd\_981&      99.09\%&97.72\%&97.94\%&&47&2162.83 ms&4243.55 ms&1601.55 mb\\
numn\_no\_35\_sd\_523&      11.24\%&10.72\%&10.88\%&&0&4491.73 ms&7025.65 ms&1601.87 mb\\
numn\_no\_50\_sd\_231&      98.81\%&97.68\%&97.74\%&&48&3495.63 ms&5577.80 ms&1601.55 mb\\
numn\_no\_50\_sd\_981&      98.85\%&97.60\%&97.86\%&&49&3590.64 ms&5648.94 ms&1601.87 mb\\
numn\_no\_50\_sd\_523&      11.24\%&10.72\%&10.88\%&&0&6423.71 ms&8945.91 ms&1601.55 mb\\
CG\_3\_sd\_231&      99.43\%&98.14\%&98.10\%&&45&1214.74 ms&3512.65 ms&1428.79 mb\\
CG\_3\_sd\_981&      99.46\%&98.02\%&98.02\%&&48&1215.65 ms&3504.93 ms&1428.79 mb\\
CG\_3\_sd\_1110&      11.24\%&11.24\%&11.84\%&&0&1207.77 ms&3485.05 ms&1428.79 mb\\
CG\_5\_sd\_231&      99.16\%&98.02\%&98.10\%&&47&1953.80 ms&4232.78 ms&1428.79 mb\\
CG\_5\_sd\_981&      99.35\%&98.08\%&98.24\%&&47&1953.24 ms&4222.54 ms&1428.79 mb\\
CG\_5\_sd\_1110&      11.24\%&11.04\%&11.64\%&&0&1942.73 ms&4204.20 ms&1428.79 mb\\
CG\_10\_sd\_231&      11.24\%&12.14\%&10.88\%&&0&3615.74 ms&5877.29 ms&1428.79 mb\\
CG\_10\_sd\_981&      9.87\%&9.22\%&10.42\%&&1&3509.07 ms&5768.25 ms&1428.79 mb\\
CG\_10\_sd\_1110&      11.24\%&12.14\%&10.88\%&&0&3782.21 ms&6046.60 ms&1428.79 mb\\
identity\_sd\_231&      99.44\%&97.90\%&98.00\%&&48&0.00 ms&2077.18 ms&839.03 mb\\
identity\_sd\_981&      99.39\%&97.98\%&98.10\%&&49&0.00 ms&2069.42 ms&839.35 mb\\
identity\_sd\_523&      99.48\%&98.14\%&98.20\%&&49&0.00 ms&2486.28 ms&839.03 mb\\
t1t2\_sd\_231&      9.92\%&9.70\%&9.84\%&&0&0.00 ms&48.85 ms&165.05 mb\\
t1t2\_sd\_981&      9.86\%&9.40\%&9.20\%&&0&0.00 ms&48.39 ms&165.05 mb\\
t1t2\_sd\_1110&      11.24\%&11.34\%&11.40\%&&0&0.00 ms&48.39 ms&165.05 mb\\
\end{tabular}
\end{table}
\endgroup

\begingroup
\begin{table}
\tiny
\centering
\caption{\textbf{CIFAR-10} - Per Weight (Full 1000 epochs).} 
\begin{tabular}{lllllllll}
\hline
  & \multicolumn{3}{c}{Accuracy} && \multicolumn{4}{c}{Training Cost}    \\ 
\cline{2-4} \cline{6-9} 
Exp   &  Training   & Validation & Test & & Convergence Epoch & Approximation Time &  Meta-Epoch Time T & Max Memory \\    \hline
numn\_no\_3\_sd\_231&      100.00\%&14.00\%&14.00\%&&41&22.97 ms&797.21 ms&875.12 mb\\
numn\_no\_3\_sd\_981&      100.00\%&92.00\%&18.00\%&&153&22.98 ms&796.71 ms&875.12 mb\\
numn\_no\_3\_sd\_1110&      100.00\%&96.00\%&22.00\%&&125&22.99 ms&797.11 ms&875.12 mb\\
numn\_no\_10\_sd\_231&      100.00\%&16.00\%&14.00\%&&14&76.66 ms&851.00 ms&875.12 mb\\
numn\_no\_10\_sd\_981&      100.00\%&12.00\%&16.00\%&&20&76.66 ms&850.85 ms&875.12 mb\\
numn\_no\_10\_sd\_1110&      100.00\%&20.00\%&16.00\%&&18&76.69 ms&850.93 ms&875.12 mb\\
numn\_no\_20\_sd\_231&      100.00\%&18.00\%&14.00\%&&57&153.38 ms&926.66 ms&875.12 mb\\
numn\_no\_20\_sd\_981&      100.00\%&18.00\%&20.00\%&&21&153.31 ms&927.02 ms&875.12 mb\\
numn\_no\_20\_sd\_1110&      100.00\%&18.00\%&20.00\%&&262&153.43 ms&928.56 ms&875.12 mb\\
numn\_no\_35\_sd\_231&      92.00\%&12.00\%&14.00\%&&0&268.53 ms&1043.49 ms&875.12 mb\\
numn\_no\_35\_sd\_981&      92.00\%&14.00\%&16.00\%&&3&268.63 ms&1043.18 ms&875.12 mb\\
numn\_no\_35\_sd\_1110&      90.00\%&10.00\%&14.00\%&&5&268.65 ms&1043.52 ms&875.12 mb\\
numn\_no\_50\_sd\_231&      10.00\%&18.00\%&8.00\%&&2&382.64 ms&1165.06 ms&875.12 mb\\
numn\_no\_50\_sd\_981&      10.00\%&18.00\%&8.00\%&&2&382.61 ms&1165.15 ms&875.12 mb\\
numn\_no\_50\_sd\_1110&      10.00\%&18.00\%&8.00\%&&2&382.23 ms&1163.97 ms&875.12 mb\\
CG\_3\_sd\_231&      100.00\%&48.00\%&20.00\%&&117&97.14 ms&870.27 ms&1101.81 mb\\
CG\_3\_sd\_981&      100.00\%&52.00\%&16.00\%&&89&97.15 ms&870.02 ms&1101.81 mb\\
CG\_3\_sd\_1110&      100.00\%&58.00\%&24.00\%&&148&97.38 ms&870.20 ms&1101.81 mb\\
CG\_5\_sd\_231&      100.00\%&58.00\%&18.00\%&&169&156.90 ms&929.38 ms&1101.81 mb\\
CG\_5\_sd\_981&      100.00\%&66.00\%&18.00\%&&172&156.51 ms&928.91 ms&1101.81 mb\\
CG\_5\_sd\_1110&      100.00\%&50.00\%&20.00\%&&137&157.10 ms&929.09 ms&1101.81 mb\\
CG\_10\_sd\_231&      2.00\%&14.00\%&4.00\%&&1&307.22 ms&1088.98 ms&1101.81 mb\\
CG\_10\_sd\_981&      2.00\%&14.00\%&4.00\%&&2&307.50 ms&1090.78 ms&1101.81 mb\\
CG\_10\_sd\_1110&      2.00\%&14.00\%&4.00\%&&1&307.26 ms&1089.58 ms&1101.81 mb\\
identity\_sd\_231&      100.00\%&76.00\%&22.00\%&&201&0.00 ms&774.13 ms&572.58 mb\\
identity\_sd\_981&      100.00\%&68.00\%&16.00\%&&41&0.00 ms&774.09 ms&572.58 mb\\
identity\_sd\_1110&      100.00\%&72.00\%&22.00\%&&154&0.00 ms&774.01 ms&572.58 mb\\
t1t2\_sd\_231&      100.00\%&66.00\%&22.00\%&&8811&0.00 ms&42.51 ms&571.97 mb\\
t1t2\_sd\_981&      100.00\%&76.00\%&22.00\%&&4838&0.00 ms&42.44 ms&571.97 mb\\
t1t2\_sd\_1110&      100.00\%&72.00\%&18.00\%&&11219&0.00 ms&42.52 ms&571.97 mb\\
\end{tabular}
\end{table}
\endgroup

\begingroup
\begin{table}
\tiny
\centering
\caption{\textbf{CIFAR-10} - Per Weight (Early Stopping).} 
\begin{tabular}{lllllllll}
\hline
  & \multicolumn{3}{c}{Accuracy} && \multicolumn{4}{c}{Training Cost}    \\ 
\cline{2-4} \cline{6-9} 
Exp   &  Training   & Validation & Test & & Convergence Epoch & Approximation Time &  Meta-Epoch Time T & Max Memory \\    \hline
numn\_no\_3\_sd\_231&      100.00\%&54.00\%&10.00\%&&41&22.88 ms&935.07 ms&875.12 mb\\
numn\_no\_3\_sd\_981&      100.00\%&100.00\%&20.00\%&&153&22.93 ms&976.70 ms&875.12 mb\\
numn\_no\_3\_sd\_1110&      100.00\%&100.00\%&26.00\%&&125&22.94 ms&976.61 ms&875.12 mb\\
numn\_no\_10\_sd\_231&      98.00\%&24.00\%&24.00\%&&14&76.56 ms&945.09 ms&875.12 mb\\
numn\_no\_10\_sd\_981&      100.00\%&26.00\%&18.00\%&&20&76.59 ms&961.41 ms&875.12 mb\\
numn\_no\_10\_sd\_1110&      100.00\%&24.00\%&18.00\%&&18&76.57 ms&961.68 ms&875.12 mb\\
numn\_no\_20\_sd\_231&      100.00\%&22.00\%&16.00\%&&57&153.18 ms&1005.33 ms&875.12 mb\\
numn\_no\_20\_sd\_981&      98.00\%&30.00\%&26.00\%&&21&153.20 ms&1048.09 ms&875.12 mb\\
numn\_no\_20\_sd\_1110&      98.00\%&14.00\%&10.00\%&&19&153.19 ms&1017.74 ms&875.12 mb\\
numn\_no\_35\_sd\_231&      88.00\%&14.00\%&14.00\%&&0&267.96 ms&1133.01 ms&875.12 mb\\
numn\_no\_35\_sd\_981&      90.00\%&12.00\%&16.00\%&&3&267.96 ms&1132.94 ms&875.12 mb\\
numn\_no\_35\_sd\_1110&      88.00\%&12.00\%&12.00\%&&5&267.96 ms&1133.05 ms&875.12 mb\\
numn\_no\_50\_sd\_231&      10.00\%&18.00\%&8.00\%&&2&381.70 ms&1364.58 ms&875.12 mb\\
numn\_no\_50\_sd\_981&      10.00\%&18.00\%&8.00\%&&2&383.59 ms&2933.08 ms&875.12 mb\\
numn\_no\_50\_sd\_1110&      10.00\%&18.00\%&8.00\%&&2&384.22 ms&3046.32 ms&875.12 mb\\
CG\_3\_sd\_231&      100.00\%&84.00\%&26.00\%&&136&97.30 ms&1032.90 ms&1101.81 mb\\
CG\_3\_sd\_981&      100.00\%&82.00\%&22.00\%&&38&97.18 ms&1014.32 ms&1101.81 mb\\
CG\_3\_sd\_1110&      100.00\%&82.00\%&22.00\%&&22&97.31 ms&1008.32 ms&1101.81 mb\\
CG\_5\_sd\_231&      100.00\%&36.00\%&24.00\%&&7&156.82 ms&991.17 ms&1101.81 mb\\
CG\_5\_sd\_981&      100.00\%&94.00\%&24.00\%&&177&157.72 ms&1034.26 ms&1101.81 mb\\
CG\_5\_sd\_1110&      100.00\%&92.00\%&26.00\%&&127&157.46 ms&1045.05 ms&1101.81 mb\\
CG\_10\_sd\_231&      12.00\%&12.00\%&10.00\%&&0&307.33 ms&1250.11 ms&1101.81 mb\\
CG\_10\_sd\_981&      12.00\%&12.00\%&10.00\%&&0&308.25 ms&2274.14 ms&1101.81 mb\\
CG\_10\_sd\_1110&      12.00\%&12.00\%&10.00\%&&0&309.70 ms&3052.81 ms&1101.81 mb\\
identity\_sd\_231&      100.00\%&86.00\%&22.00\%&&75&0.00 ms&912.48 ms&572.58 mb\\
identity\_sd\_981&      100.00\%&76.00\%&20.00\%&&41&0.00 ms&900.30 ms&572.58 mb\\
identity\_sd\_1110&      100.00\%&80.00\%&22.00\%&&100&0.00 ms&2162.10 ms&572.58 mb\\
\end{tabular}
\end{table}
\endgroup

\begingroup
\begin{table}
\tiny
\centering
\caption{\textbf{CIFAR-10} - Per Layer.} 
\begin{tabular}{lllllllll}
\hline
  & \multicolumn{3}{c}{Accuracy} && \multicolumn{4}{c}{Training Cost}    \\ 
\cline{2-4} \cline{6-9} 
Exp   &  Training   & Validation & Test & & Convergence Epoch & Approximation Time &  Meta-Epoch Time T & Max Memory \\    \hline
numn\_no\_3\_sd\_231&      98.67\%&55.58\%&55.08\%&&28&3101.19 ms&12984.61 ms&5601.64 mb\\
numn\_no\_3\_sd\_981&      98.93\%&55.66\%&55.10\%&&25&3100.71 ms&12995.64 ms&5601.64 mb\\
numn\_no\_3\_sd\_1110&      98.32\%&55.64\%&55.18\%&&38&3101.03 ms&12994.88 ms&5601.64 mb\\
numn\_no\_10\_sd\_231&      31.41\%&32.08\%&30.50\%&&43&10356.82 ms&20225.30 ms&5601.64 mb\\
numn\_no\_10\_sd\_981&      31.54\%&32.00\%&31.06\%&&5&10352.13 ms&20243.04 ms&5601.64 mb\\
numn\_no\_10\_sd\_1110&      41.10\%&41.26\%&40.82\%&&48&10360.92 ms&20293.44 ms&5601.64 mb\\
numn\_no\_20\_sd\_231&      27.55\%&28.84\%&23.32\%&&2&20758.27 ms&30623.70 ms&5601.64 mb\\
numn\_no\_20\_sd\_981&      30.98\%&31.72\%&30.42\%&&5&20737.55 ms&30662.66 ms&5601.64 mb\\
numn\_no\_20\_sd\_1110&      33.86\%&33.92\%&31.44\%&&9&20743.40 ms&30628.73 ms&5601.64 mb\\
numn\_no\_35\_sd\_231&      29.93\%&30.82\%&22.98\%&&43&36252.12 ms&46182.87 ms&5601.64 mb\\
numn\_no\_35\_sd\_981&      27.93\%&29.24\%&23.06\%&&4&36254.22 ms&46170.08 ms&5601.64 mb\\
numn\_no\_35\_sd\_1110&      27.83\%&29.24\%&24.20\%&&3&36358.74 ms&46240.10 ms&5601.64 mb\\
numn\_no\_50\_sd\_231&      25.34\%&26.70\%&23.22\%&&3&51963.54 ms&61859.36 ms&5601.64 mb\\
numn\_no\_50\_sd\_981&      27.78\%&28.76\%&23.30\%&&4&51729.27 ms&61616.66 ms&5601.64 mb\\
numn\_no\_50\_sd\_1110&      30.53\%&31.06\%&25.12\%&&5&51956.86 ms&61910.07 ms&5601.64 mb\\
CG\_3\_sd\_231&      98.96\%&56.80\%&54.08\%&&28&9654.95 ms&18798.29 ms&5214.49 mb\\
CG\_3\_sd\_981&      98.43\%&56.30\%&55.26\%&&37&9658.85 ms&18804.82 ms&5214.49 mb\\
CG\_3\_sd\_1110&      98.91\%&56.10\%&55.16\%&&44&9644.06 ms&18761.14 ms&5214.49 mb\\
CG\_5\_sd\_231&      34.84\%&35.92\%&34.76\%&&49&15442.94 ms&24572.34 ms&5214.49 mb\\
CG\_5\_sd\_981&      40.69\%&41.42\%&39.80\%&&46&15435.95 ms&24561.46 ms&5214.49 mb\\
CG\_5\_sd\_1110&      31.23\%&31.60\%&10.32\%&&2&13863.82 ms&22843.40 ms&5214.49 mb\\
CG\_10\_sd\_231&      16.76\%&16.58\%&10.16\%&&0&26809.23 ms&35739.90 ms&5214.49 mb\\
CG\_10\_sd\_981&      17.21\%&16.90\%&10.16\%&&0&26880.70 ms&35814.25 ms&5214.49 mb\\
CG\_10\_sd\_1110&      21.52\%&22.22\%&10.16\%&&1&26843.49 ms&35799.44 ms&5214.49 mb\\
identity\_sd\_231&      98.19\%&55.74\%&53.92\%&&37&0.00 ms&9931.71 ms&3029.88 mb\\
identity\_sd\_981&      99.13\%&55.88\%&55.88\%&&49&0.00 ms&9870.14 ms&3029.88 mb\\
identity\_sd\_1110&      98.45\%&55.44\%&55.26\%&&39&0.00 ms&9767.52 ms&3029.88 mb\\
t1t2\_sd\_231&      99.93\%&56.34\%&55.34\%&&9461&0.00 ms&321.14 ms&2415.42 mb\\
t1t2\_sd\_981&      99.89\%&56.18\%&54.70\%&&22069&0.00 ms&321.19 ms&2415.42 mb\\
t1t2\_sd\_1110&      99.91\%&56.70\%&56.08\%&&14109&0.00 ms&321.15 ms&2415.42 mb\\
\end{tabular}
\end{table}
\endgroup

\clearpage
\section{Baselines Regularization Results}\label{app:baselines}
\begingroup
\begin{table}[h]
\centering
\caption{Baseline results for various regularization weights for an MLP with as many hidden neurons as inputs.} 
\begin{tabular}{cccc}
\hline
  & \multicolumn{3}{c}{Accuracy}    \\ 
\cline{2-4}
$L_2$ Weight   &  Training   & Validation & Test \\
\hline
\multicolumn{4}{c}{\textbf{CIFAR-10}}\\
\hline
$1\times10^{-6}$&58.29$\pm$0.24\%&51.60$\pm$0.20\%&50.39$\pm$0.54\%\\
$1\times10^{-5}$&58.32$\pm$0.24\%&51.64$\pm$0.21\%&50.41$\pm$0.69\%\\
$1\times10^{-4}$&58.17$\pm$0.20\%&51.54$\pm$0.23\%&50.39$\pm$0.56\%\\
$1\times10^{-3}$&57.57$\pm$0.21\%&51.39$\pm$0.21\%&50.09$\pm$0.76\%\\
$1\times10^{-2}$&53.09$\pm$0.30\%&49.70$\pm$0.31\%&47.81$\pm$0.43\%\\
$1\times10^{-1}$&42.29$\pm$0.23\%&42.31$\pm$0.49\%&40.82$\pm$0.88\%\\
$5\times10^{-1}$&34.48$\pm$0.18\%&34.94$\pm$0.18\%&29.93$\pm$0.48\%\\
1&31.06$\pm$0.46\%&31.64$\pm$0.31\%&22.43$\pm$0.34\%\\
2&28.24$\pm$0.40\%&28.67$\pm$0.26\%&15.55$\pm$0.43\%\\
3&26.28$\pm$0.83\%&26.52$\pm$0.53\%&10.05$\pm$0.06\%\\
\hline
\multicolumn{4}{c}{\textbf{MNIST}}\\
\hline
$1\times10^{-6}$&94.78$\pm$0.10\%&94.81$\pm$0.25\%&94.41$\pm$0.14\%\\
$1\times10^{-5}$&94.78$\pm$0.09\%&94.82$\pm$0.24\%&94.41$\pm$0.14\%\\
$1\times10^{-4}$&94.78$\pm$0.09\%&94.82$\pm$0.25\%&94.43$\pm$0.12\%\\
$1\times10^{-3}$&94.73$\pm$0.10\%&94.81$\pm$0.23\%&94.39$\pm$0.17\%\\
$1\times10^{-2}$&93.99$\pm$0.10\%&94.28$\pm$0.26\%&93.67$\pm$0.09\%\\
$1\times10^{-1}$&89.89$\pm$0.06\%&90.72$\pm$0.10\%&89.93$\pm$0.21\%\\
$5\times10^{-1}$&82.24$\pm$0.30\%&83.48$\pm$0.13\%&80.39$\pm$0.52\%\\
1&74.26$\pm$0.25\%&75.80$\pm$0.28\%&55.37$\pm$1.56\%\\
2&67.67$\pm$2.16\%&69.13$\pm$1.64\%&15.57$\pm$3.35\%\\
3&57.75$\pm$3.88\%&59.15$\pm$3.36\%&11.17$\pm$0.15\%\\
\end{tabular}
\end{table}
\endgroup

\clearpage
\section{Training Evolution Plots for Main Experiments}
\begin{figure}[h]
\begin{center}
\centerline{\includegraphics[width=1\columnwidth]{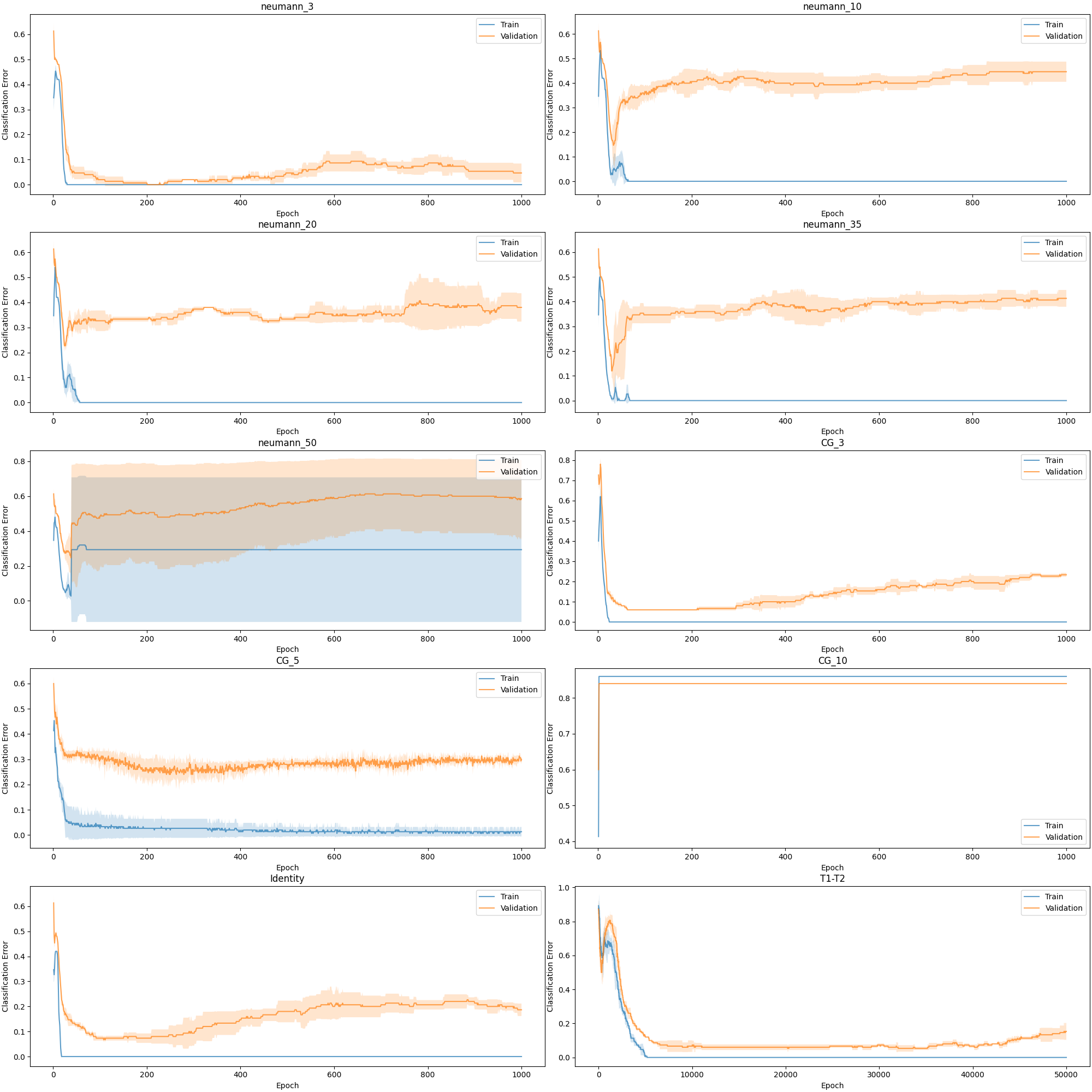}}
\caption{Training Metrics Evolution for MNIST Overfitting a Small Validation Set Experiments.}
\end{center}
\end{figure}

\begin{figure}[tb]
\begin{center}
\centerline{\includegraphics[width=1\columnwidth]{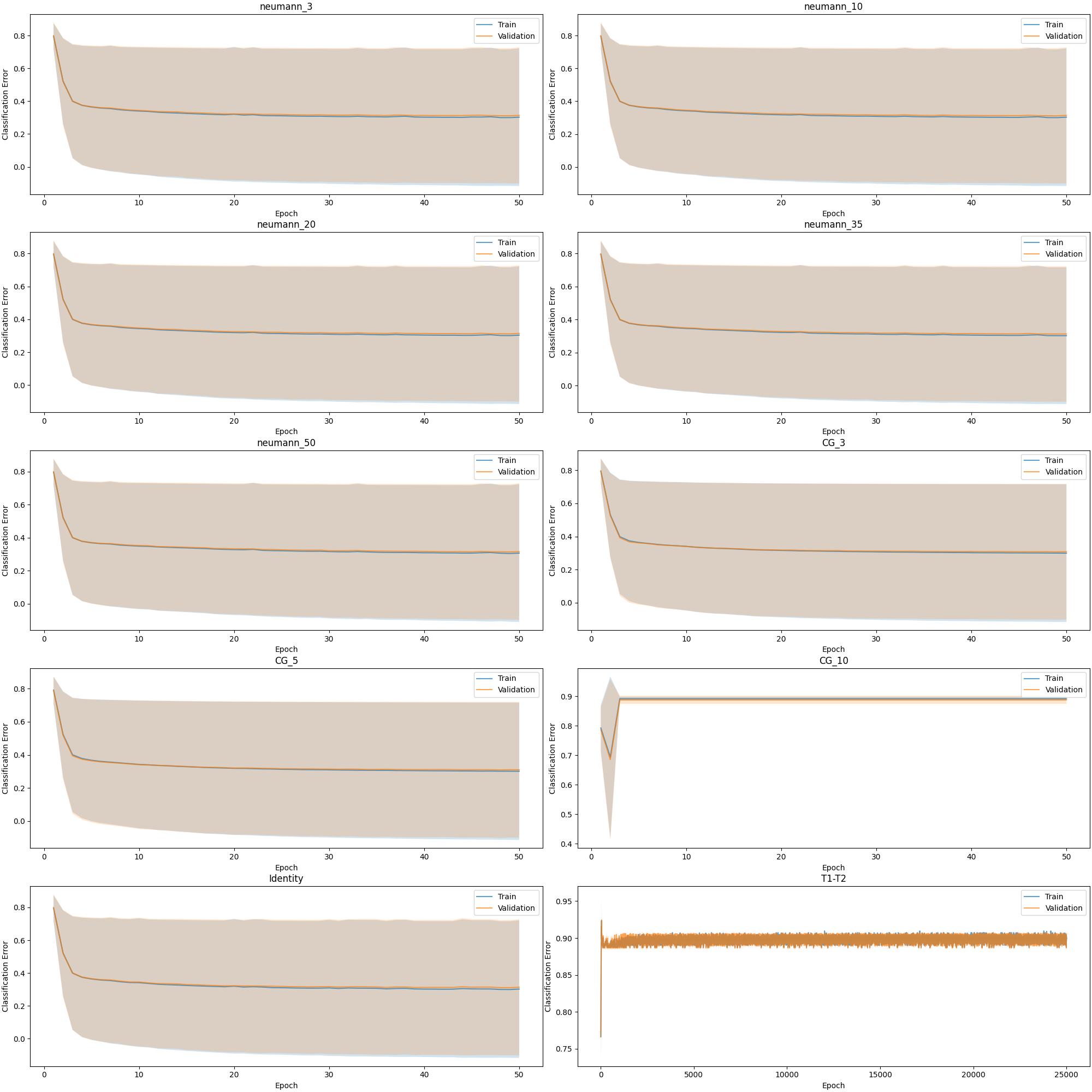}}
\caption{Training Metrics Evolution for MNIST Learning Per Layer $L_2$ Regularization Weight Experiments. Note the extreme high variance and bad performance of most methods here. This is a result of including bad seeds whose training was unstable. As stated, the results in the main body of the report only included stable seeds since finding working seeds was challenging.}
\end{center}
\end{figure}

\begin{figure}[tb]
\begin{center}
\centerline{\includegraphics[width=1\columnwidth]{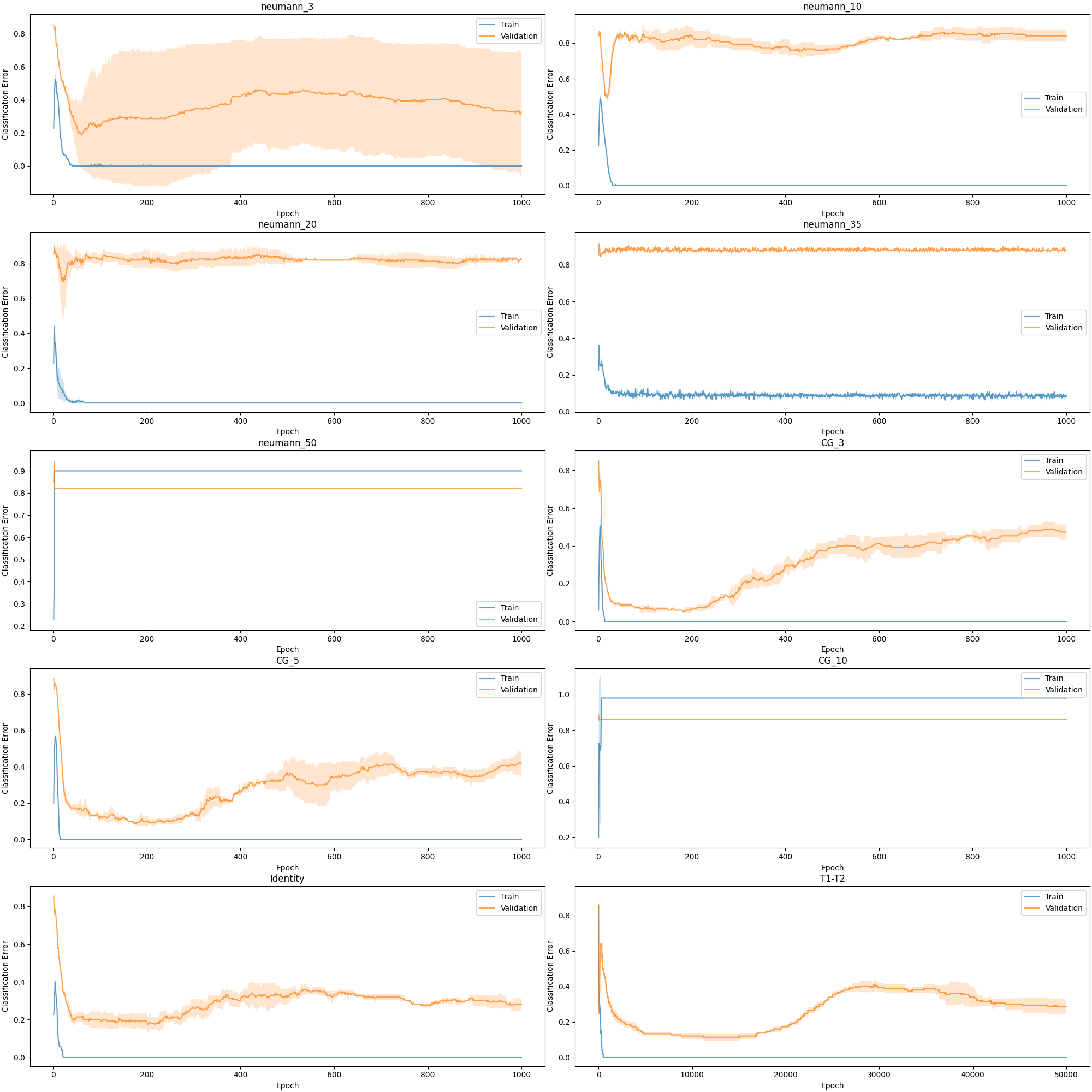}}
\caption{Training Metrics Evolution for CIFAR-10 Overfitting a Small Validation Set Experiments.}
\end{center}
\end{figure}

\begin{figure}[tb]
\begin{center}
\centerline{\includegraphics[width=1\columnwidth]{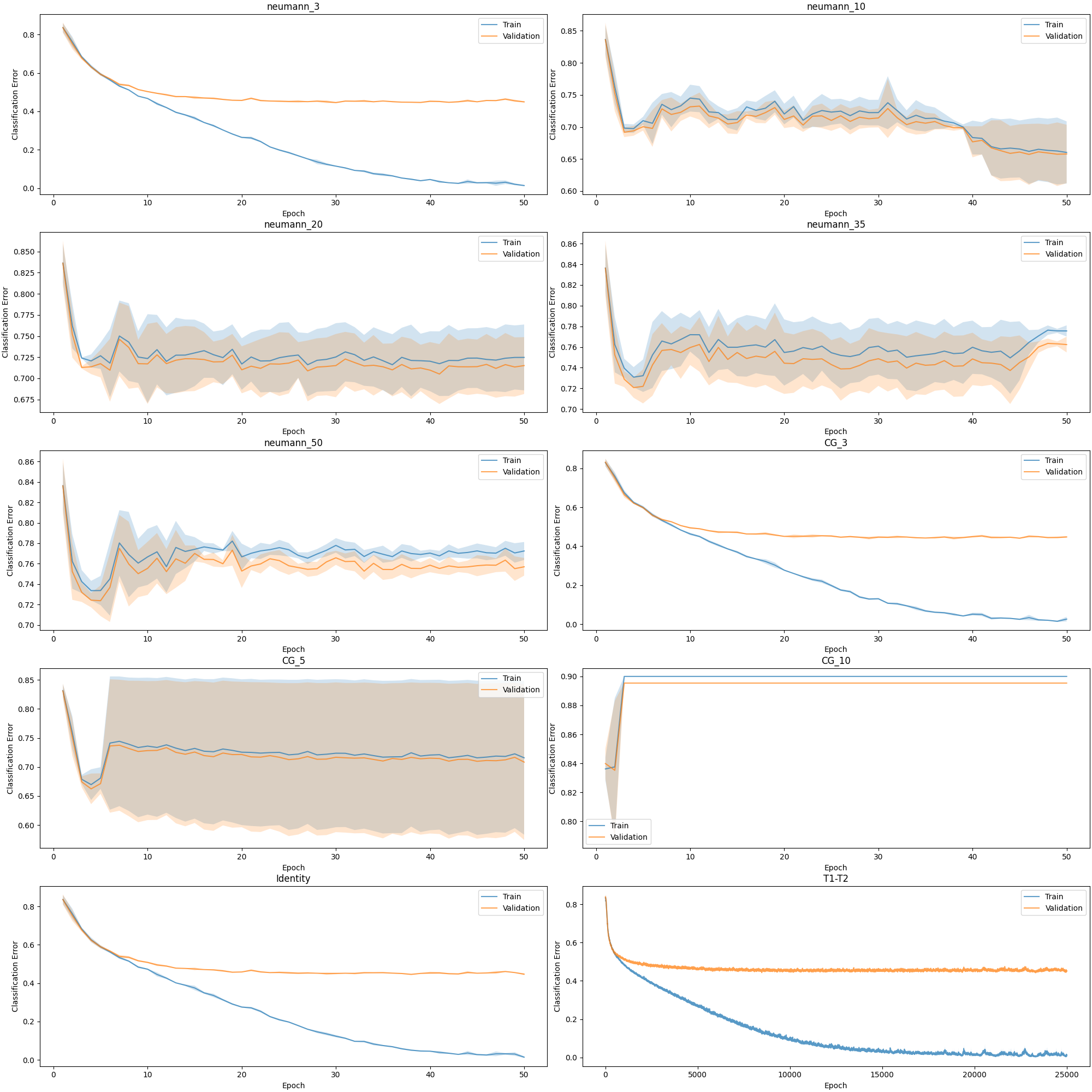}}
\caption{Training Metrics Evolution for CIFAR-10 Learning Per Layer $L_2$ Regularization Weight Experiments.}
\end{center}
\end{figure}

%% file: 99.appendix_ablation.tex
\chapter{Ablation Study Training Evolution Plots}\label{app:ablation_evolution_plots}
\begin{figure}[h]
\begin{center}
\centerline{\includegraphics[width=1\columnwidth]{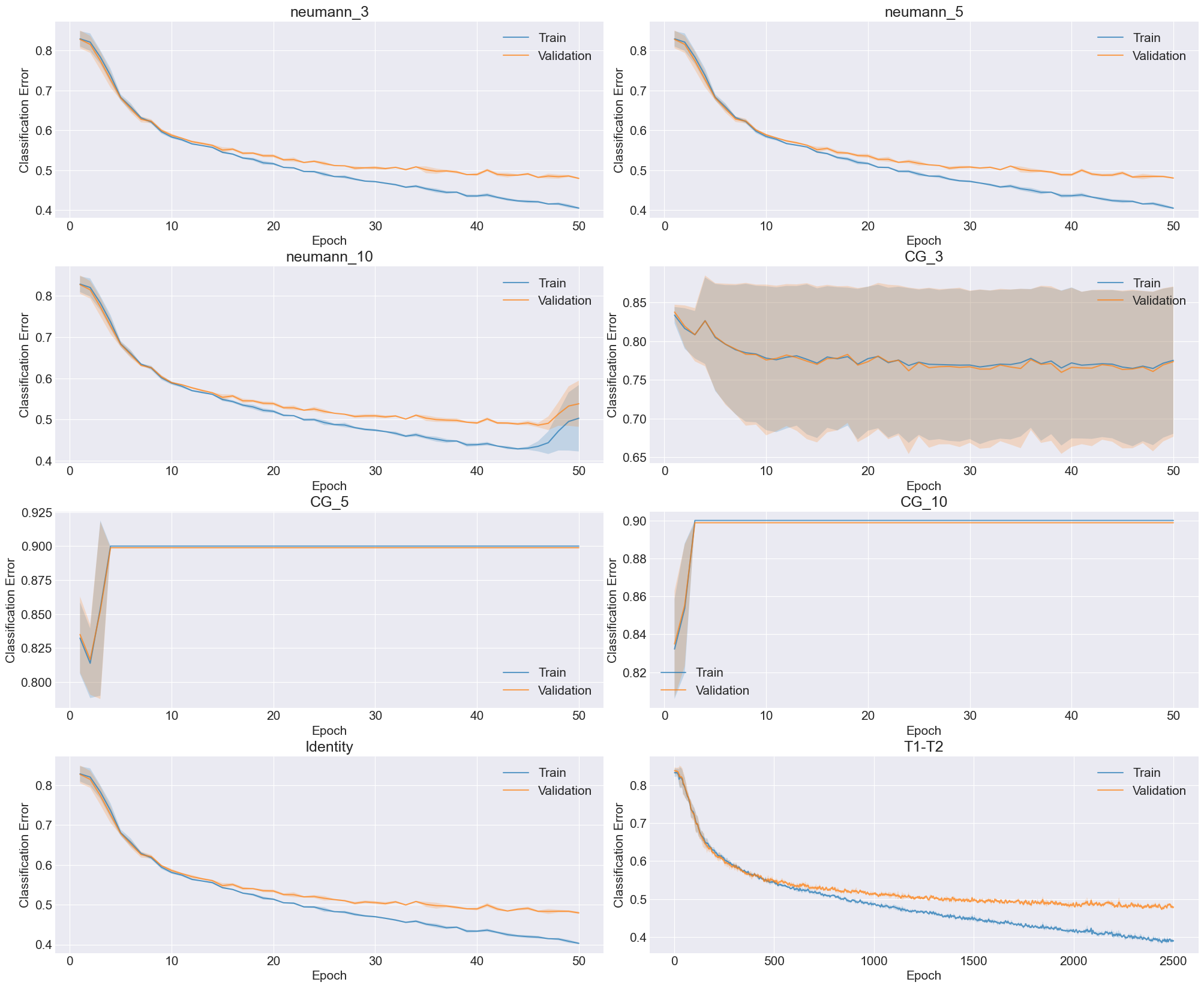}}
\caption{Training Metrics Evolution for Ablation Study, section \ref{sec:ablation}, when using 50 inner loop steps.}
\end{center}
\end{figure}

\begin{figure}[tb]
\begin{center}
\centerline{\includegraphics[width=1\columnwidth]{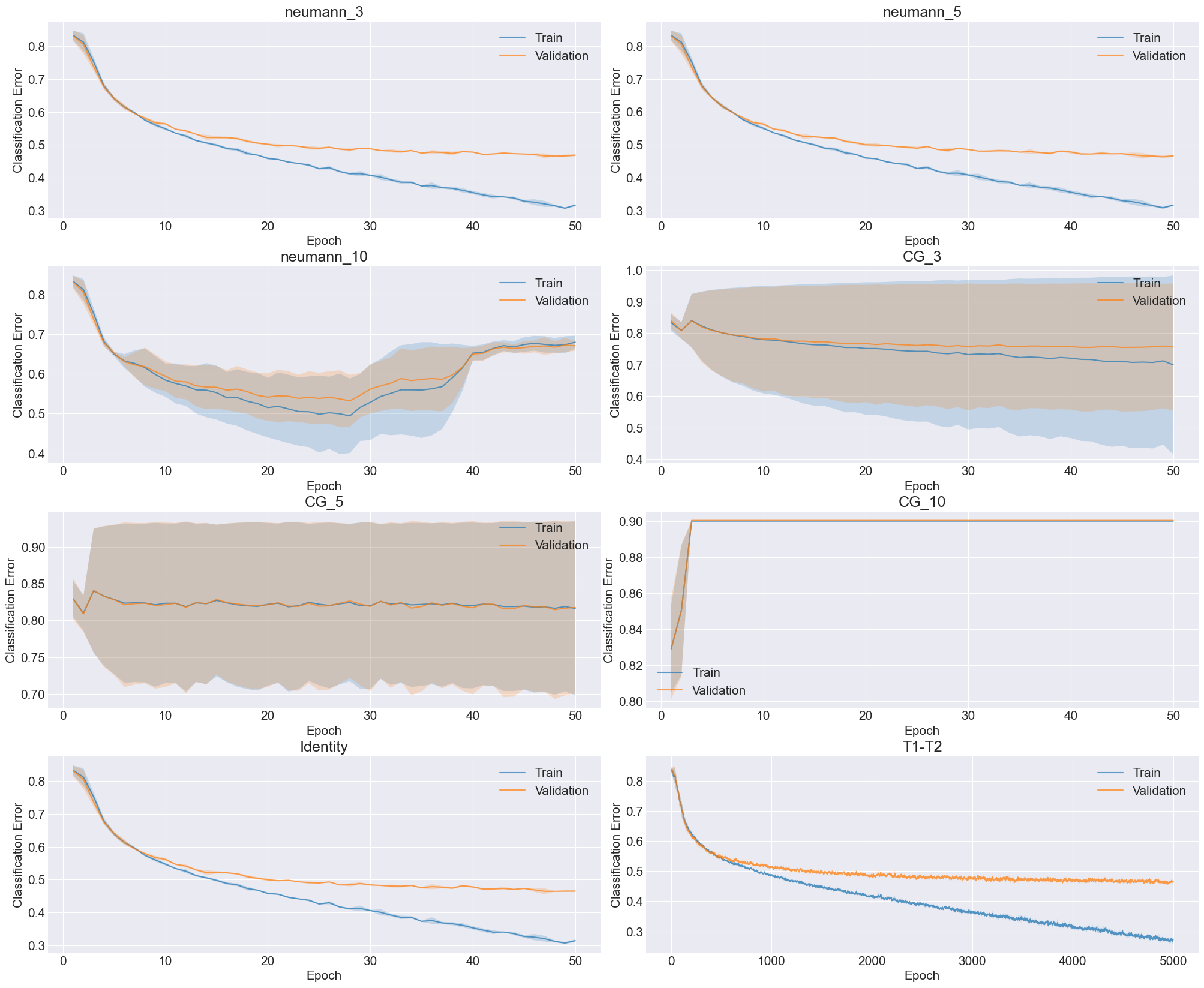}}
\caption{Training Metrics Evolution for Ablation Study, section \ref{sec:ablation}, when using 100 inner loop steps.}
\end{center}
\end{figure}

\begin{figure}[tb]
\begin{center}
\centerline{\includegraphics[width=1\columnwidth]{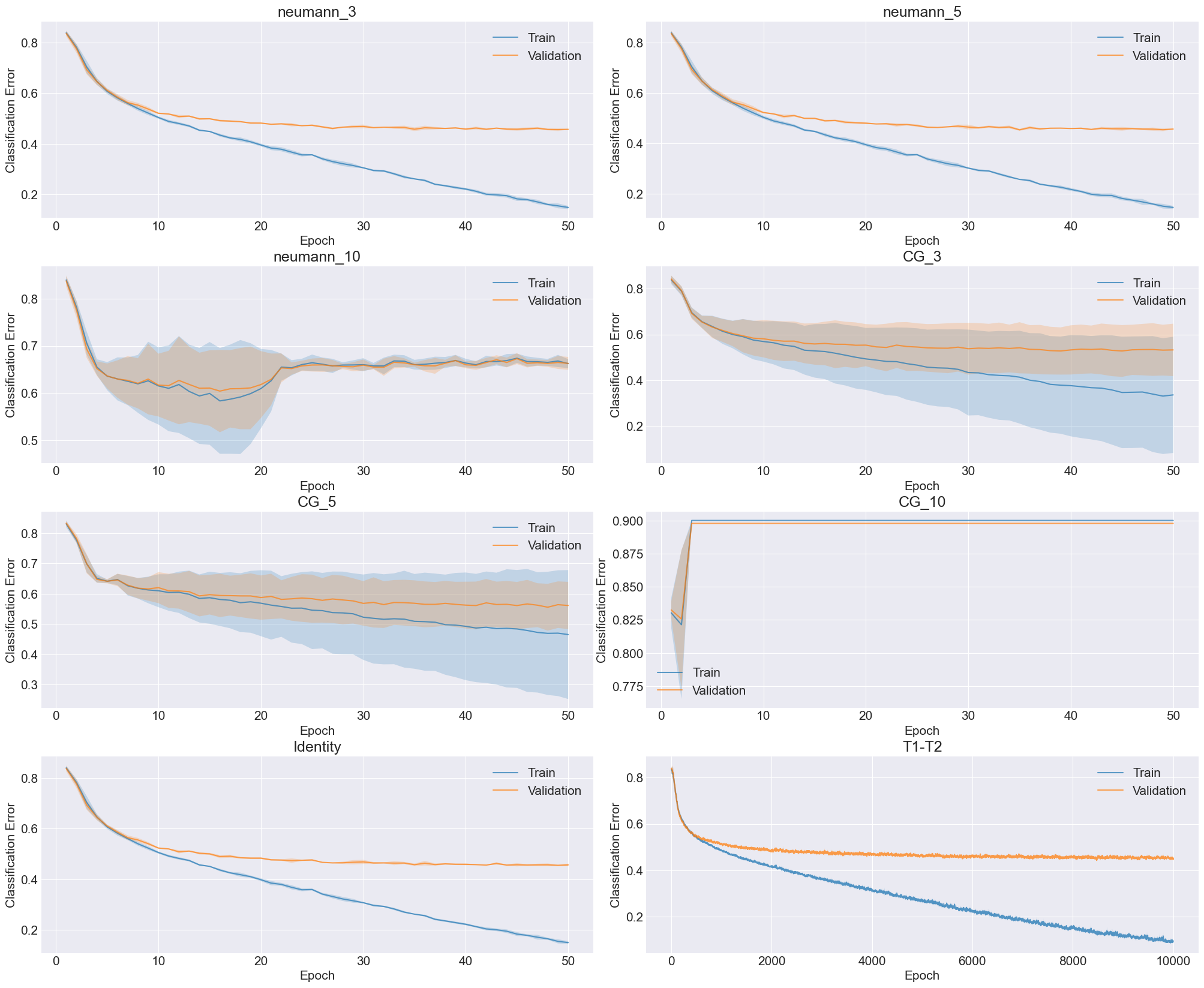}}
\caption{Training Metrics Evolution for Ablation Study, section \ref{sec:ablation}, when using 200 inner loop steps.}
\end{center}
\end{figure}

\begin{figure}[tb]
\begin{center}
\centerline{\includegraphics[width=1\columnwidth]{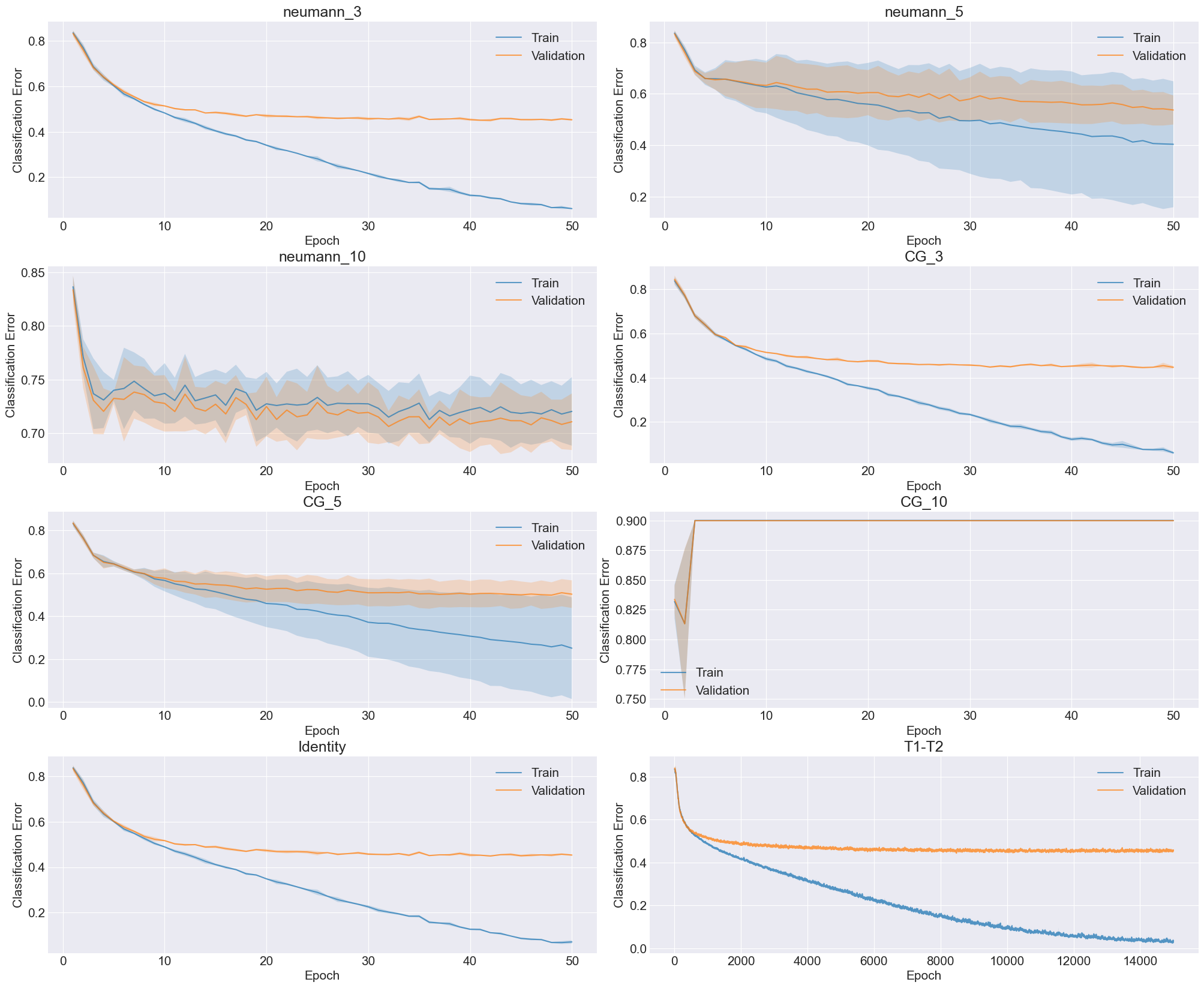}}
\caption{Training Metrics Evolution for Ablation Study, section \ref{sec:ablation}, when using 300 inner loop steps.}
\end{center}
\end{figure}

\begin{figure}[tb]
\begin{center}
\centerline{\includegraphics[width=1\columnwidth]{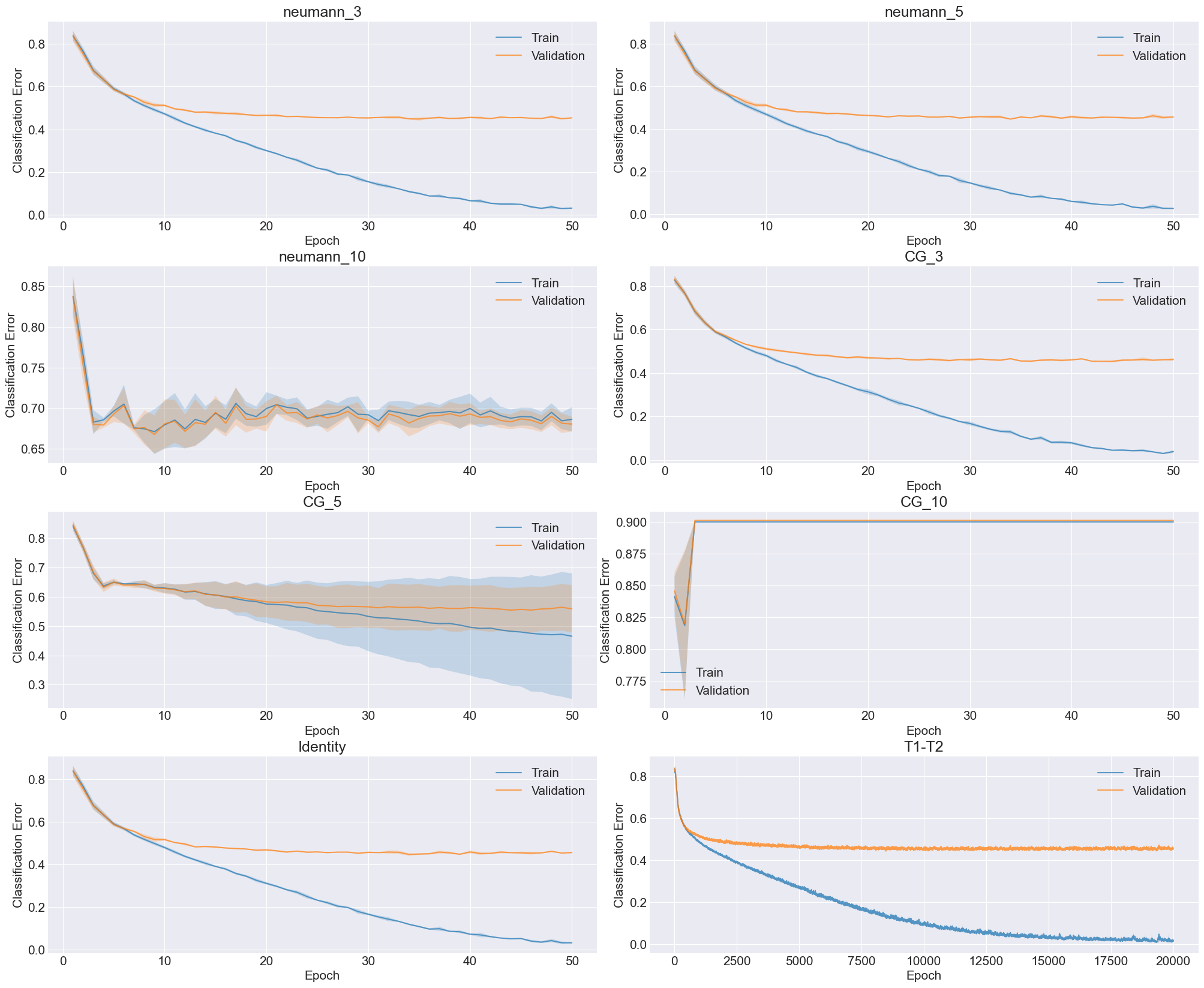}}
\caption{Training Metrics Evolution for Ablation Study, section \ref{sec:ablation}, when using 400 inner loop steps.}
\end{center}
\end{figure}

\begin{figure}[tb]
\begin{center}
\centerline{\includegraphics[width=1\columnwidth]{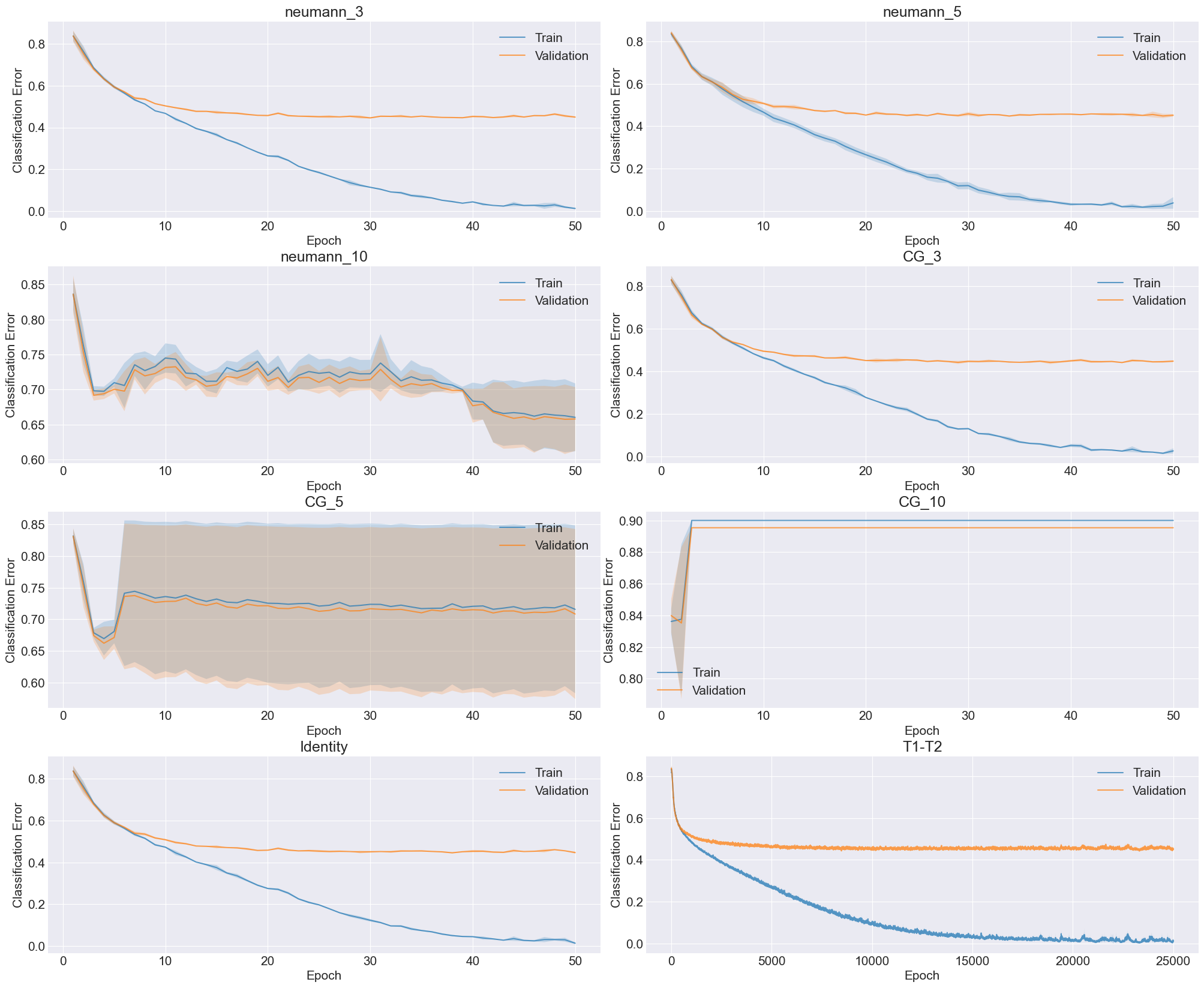}}
\caption{Training Metrics Evolution for Ablation Study, section \ref{sec:ablation}, when using 500 inner loop steps.}
\end{center}
\end{figure}